\documentclass[10pt,journal,compsoc]{IEEEtran}

\usepackage{subfig}
\usepackage{amssymb}
\usepackage{amsfonts}
\usepackage{bm}
\usepackage{mathrsfs}
\usepackage{mathtools}
\usepackage{outlines}
\usepackage{import}
\usepackage{color, soul} 
\usepackage{pdflscape}

\definecolor{myred3}{RGB}{235,35,44}

\ifCLASSOPTIONcompsoc
  \usepackage[nocompress]{cite}
\else
  \usepackage{cite}
\fi

\ifCLASSINFOpdf
  \usepackage[pdftex]{graphicx}
  \graphicspath{{./images/}}
\else
\fi

\begin{document}

\title{Neural Network Layer Algebra: A Framework to Measure Capacity and Compression in Deep Learning}

\author{Alberto~Badias
        and~Ashis~G.~Banerjee,~\IEEEmembership{Senior Member,~IEEE}
\thanks{A. Badias is with the Technical University of Madrid (UPM), Spain, and was a visiting scholar in the Department of Mechanical Engineering, University of Washington, Seattle WA 98195, USA; e-mail: alberto.badias@upm.es.}
\thanks{A. G. Banerjee is with the Department of Industrial \& Systems Engineering and the Department of Mechanical Engineering, University of Washington, Seattle WA 98195, USA; e-mail: ashisb@uw.edu.}}

\IEEEtitleabstractindextext{%
\begin{abstract}
We present a new framework to measure the intrinsic properties of (deep) neural networks. While we focus on convolutional networks, our framework can be extrapolated to any network architecture. In particular, we evaluate two network properties, namely, \emph{capacity}, which is related to \emph{expressivity}, and \emph{compression}, { which is related to \emph{learnability}}. Both these properties depend only on the network structure and are independent of the network parameters. To this end, we propose two metrics: the first one, called \emph{layer complexity}, captures the architectural complexity of any network layer; and, the second one, called \emph{layer intrinsic power}, encodes how data is compressed along the network. The metrics are based on the concept of \emph{layer algebra}, which is also introduced in this paper. This concept is based on the idea that the global properties depend on the network topology, and the leaf nodes of any neural network can be approximated using local transfer functions, thereby, allowing a simple computation of the global metrics. { We show that our global complexity metric can be calculated and represented more conveniently than the widely-used VC dimension.} We also compare the properties of various state-of-the art architectures using our metrics and use the properties to analyze their accuracy on benchmark image classification datasets.
\end{abstract}

}

\maketitle

\IEEEdisplaynontitleabstractindextext

\IEEEpeerreviewmaketitle

\section{Introduction}\label{sec:introduction}

\IEEEPARstart{D}{eep} learning models have a good ability to deal with challenging problems that are too complex for us to explain by means of simple and deterministic laws in closed forms. Some examples include the extraction of relevant information from images \cite{Krizhevsky:2012aa}, image inpainting and denoising \cite{Xie:2012aa}, natural language processing \cite{Collobert:2008aa}, the creation of music \cite{Carr:2018aa}, and learning how to play a 3D role-playing game properly \cite{openAIDota}.

In particular, if we consider the image processing problem, deep learning methods deal with data in their natural forms (images) and try to make the system capable of automatically extracting semantic information. In other words, we are not only dealing with the problem of prediction (here, classification), but also of discovering the representations needed to extract discriminative features from raw data. It is, therefore, not only a question of obtaining a target function capable of classifying data that is already in a defined space, but also a question of defining that space simultaneously. 

The latter question is difficult to answer since the true dimension of the problem space is usually unknown and depends on both the input and output data. For example, the number of classes (labels) is of importance to the classifier. 
Experience tells us that some of the network structures are focused on the extraction of relevant descriptors, such as convolutions for images and recurrent units for time series data, while other structures have a direct prediction-related objective, such as dense groups of fully connected layers. A deep neural network for any challenging task usually includes both the types of structures, creating a complex architecture of information flow channels. Since we have no \textit{a priori} idea about the dimensionality of the problem space, we usually tend to create a large set of interconnected layers to achieve, as best as possible, high accuracy in solving a given task. Although it is true that people try to reduce the network size once the desired accuracy is achieved, they typically do not aim to attain any control based on the problem dimensionality.

For the moment, we think there are not many ways to estimate how big, small or oversized a network is. In fact, to the best of our knowledge, there is no tool to compare the different architectures systematically (for example, we think that the sum of the parameters is not a representative comparison measure).
 {While approximation and generalization errors are widely used to evaluate supervised learning models, we base our evaluation on two related but slightly different metrics of expressivity and learnability \cite{Rolnick:2018aa}, as they pertain directly to the ability of networks to encode complex functions and identify the underlying patterns in the input data. Accordingly, we draw inspiration from the capacity and compression of (data) flow in networks to define two corresponding metrics, complexity and intrinsic power, which are related to expressivity and learnability, respectively. We first define our metrics on the individual network layers and then combine them to obtain (scalar) global cumulative values based only on the network topology. We show that these values are practically computable for extremely large networks, and are useful in estimating the ease of training and explaining prediction accuracy on benchmark image classification problems.}

The rest of this paper is organized as follows. 
Section 2 summarizes the related work on the other methods that are used to measure certain properties of neural networks. Section 3 presents an overview of our method. In Section 4, the net properties and metrics are introduced. Section 5 explains the evaluation of these metrics for a set of known local layers. Section 6 defines the concept of layer algebra, a necessary tool to take into account the topology of the architecture into the metrics. Section 7 describes the experiments carried out to assess the usefulness of the metrics. Finally, Section 8 provides some concluding remarks and outlines future work directions.

\section{Related Work}

Although there has been a lot of interest in neural networks in the last decade, largely due to the computational power of the modern devices, the theoretical principles were developed earlier. For example, we know that a neural network with a hidden layer is a universal approximator, which can approximate any continuous function on compact subsets of $\mathbb{R}^n$ arbitrarily well, under some assumptions on the activation functions \cite{Hanin:2019aa}. However, some aspects of neural networks, especially pertaining to deep architectures, are unknown to the scientific world.

There are many works that define the basic properties of neural networks and propose methods to estimate these properties. To this end, some researchers demonstrated properties such as learning ability, generalization, and robustness \cite{Vidal:2017aa, Rolnick:2018aa}. Baldi and Vershynin \cite{Baldi:2018aa} defined the capacity of a network as the binary logarithm of the number of the functions it can implement. 

A lot of attention has been given on estimating network {\em complexity} in particular, since it is one of the most interesting but difficult-to-measure properties. An example is the development of a complexity theory based on the computability of functions by neural networks of a given type and size \cite{Orponen:1994aa}. Subsequently, Montufar \textit{et al.} \cite{Montufar:2014aa} estimated the complexity of functions that are computable by deep feedforward neural networks with linear activation functions by observing the number of generated linear regions. Others proved that an upper bound to the number of linear regions scales exponentially with network depth but polynomially with width \cite{Pascanu:2013aa, Novak:2018aa, Hu:2020aa}, while some works notice that size and depth of the networks affect the ability to approximate real functions \cite{Vardi:2021aa}. Song \textit{et al.} \cite{Song:2017aa} derived a theoretical lower bound to imply that efficient training algorithms require stronger assumptions on the target functions and input data distributions than Lipschitz continuity and smoothness. 

Cohen \textit{et al.} \cite{Cohen:2016aa} analyzed the complexity of the functions that can be expressed by a network using tensor analysis, thereby, comparing deep and shallow architectures using factors such as locality, sharing, and pooling to establish an equivalence between the networks and hierarchical tensor factorization. Raghu \textit{et al.} focused on overall network expressivity, and measured it as the length of the trajectory that captures the change in the output as the input sweeps along a one-dimensional path \cite{Raghu:2017aa}. They proved that expressivity complexity grows exponentially with network depth \cite{Raghu:2016aa}. 

Another way of estimating the complexity of network is by means of its sensitivity to input perturbations. Accordingly, Novak \textit{et al.} \cite{Novak:2018aa} distinguished different network models based on their sensitivity and came up with a robustness measure through their input-output Jacobian. Alternatively, several researchers have connected neural networks to dynamical models to estimate the complexity from a theoretical perspective by establishing a parallel between the network architectures and stochastic partial differential equations \cite{Goldt:2017aa, Han:2018aa, Sun:2018aa}.
A different approach, like Zhang \textit{et al.} \cite{Zhang:2016aa}, is focused on complexity estimation of recurrent neural networks by using the concept of cyclic graph to define recurrent depth or recurrent skip coefficient to capture how rapidly information propagates over time. 

Other representative efforts include the work by Bartlett and Mendelson \cite{Bartlett:2002aa} to compute the complexity of a classification function using the Rademacher and Gaussian complexities. Bianchini and Scarselli \cite{Bianchini:2014aa} used the topology of the layers (specifically, Betti number) to estimate the complexity of the network. On a related note, Rieck \textit{et al.} developed a topological complexity measure, called neural persistence, based on the use of persistent homology \cite{Rieck:2019aa}. Barron and Klusowski \cite{Barron:2018aa} determined the theoretical accuracy by measuring the statistical risk as defined by the mean squared prediction error. They also proposed methods to estimate metric entropy, in addition to complexity and statistical risk, in \cite{Barron:2019aa}. 

From a more practical standpoint, it is common to quantify network complexity using the number of training parameters, the cost of memory storage, or the total number of operations required (FLOPS) to evaluate the data during inference \cite{Canziani:2016aa}. Another simple way to compare the complexity of neural networks is by empirically measuring the time taken by a computer to perform the complete computation. However, as might be expected, this depends on the computer being used, the data set, and the implementation itself. Other approaches directly measure the complexity based on the depth of the network as given by the number of layers. We believe that this is not an appropriate metric since a layer can be very complex, involving many operations, or it can be very simple, with only an activation function. In fact, Zhang \textit{et al.} \cite{Zhang:2016aa} demonstrated interesting performance characteristics of recurrent neural networks by defining depth as the maximum number of nonlinear transformations from the inputs to the outputs.

Beyond property estimation, researchers have compared deep architectures with shallow architectures, noting that deeper networks have better learning capabilities for really challenging problems \cite{Rolnick:2017aa}. They have proved that deep networks can approximate the class of compositional functions with the same accuracy as shallow networks but with an exponentially lower number of training parameters \cite{Mhaskar:2016aa}, as well as VC-dimension (Vapnik–Chervonenkis) \cite{Bartlett:2003aa}. They have also tried to answer how width affects the expressiveness of neural networks, thereby, obtaining a universal approximation theorem for width-bounded ReLU networks \cite{Lu:2017aa}.

\section{Overview of the Proposed Method}

\begin{figure}[!h]
	\centering
	\small{
	\includegraphics[width=0.49\textwidth]{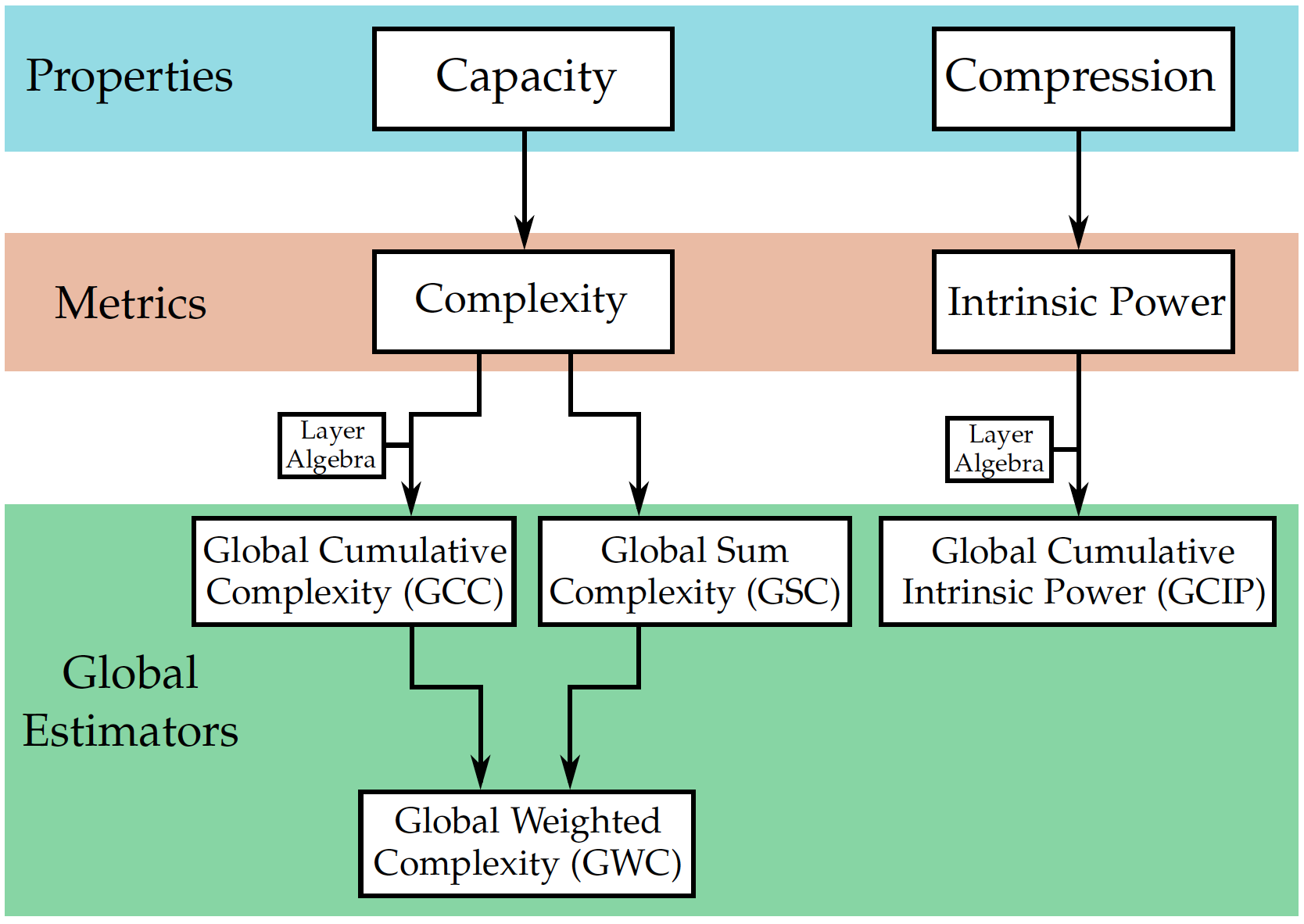}}
	\caption{Schematic overview of the proposed concepts.}
	\label{fig:Overview}
\end{figure}

As we have seen, several methods have been devised to estimate the complexity of neural networks. Here, we propose an alternate method that measures the properties of a network architecture precisely using specific values rather than limits or bounds. On many occasions, we talk about deep networks, and different architectures are compared based on the number of layers. However, we think this is not the best approach since, in addition to the number of layers, it is necessary to analyze how each (local) layer affects the global complexity. 
Similarly, when the layers have large widths, it is possible for a network to be shallow but have many parallel channels that enable much greater complexity than a deeper network. Therefore, we believe that a rigorous mathematical framework is necessary to standardize neural network architectures and allow fair comparisons among the different architectures.

Our goal is to establish a methodology to compare the different architectures encompassing the concepts of depth, width or resolution \cite{Tan2019}. To do so, we present a set of metrics to estimate the local values of our properties in each layer (see Fig.~\ref{fig:Overview}). Subsequently, we apply the notion of \emph{layer algebra} to obtain the global values. We propose two properties, capacity and compression, for which we use two metrics: the first measures the \emph{layer complexity} of a neural network and the second measures the \emph{intrinsic power} 
of the data.  {Note that our metrics are related to the expressivity and learnability properties established in \cite{Rolnick:2018aa}, which affect the approximation and generalization errors on supervised learning problems. }

According to our hypotheses, two networks with the same architecture but different parameters, would obtain the same values of our proposed metrics. This happens because our measurements are only based on the operations or transformations that the data undergoes throughout the network.
While we think it is better to use analytical functions to estimate our metrics, 
we rely on a data-driven approach to compute the values for certain network layers that use activation functions. We believe that our method could be used to predict if a network would be accurate enough, learnable, or computationally intensive before actually training or testing the network on large-scale data sets.

\section{Net Properties}
\label{Sec:NetProperties}

 {In this work, we focus on the steady-state properties of a neural network.} We are not going to model the learning process from the transitory viewpoint of optimizing a set of parameters such as weights and biases. To do this, we look at the changes that the data undergoes as it flows through the network based on its morphology. 

Let us suppose we have a data set $X \in \mathbb{R}^{N \times I}$ with $N$ as the number of samples and $I$ as the dimension of the input data. Let $Y \in \mathbb{R}^{N \times C}$ be the output data set, with $C$ being the size of the output data. $C$ is a continuous set if we are facing a regression problem or categorical if it is a classification problem. The changes applied by a neural network on the input data to obtain the output (assuming supervised learning) can be defined as a mapping function $f(X): X \rightarrow Y$. Depending on the problem complexity, the function $f(X)$ may need to be highly non-linear, where this global function is divided into a set of $L$ compositional local functions $f^{l}(X)$ that are linked in a multiplicative way to recover the function $f(X)$. Note that $L$ is often referred as the number of layers. For example, in the image classification context, $f^l$ 
comprises both linear transformations and component-wise non-linear functions, also called activation functions. $f^l$ can then be expressed in a matrix form as $f^{l}(x) = \sigma (W^{l} x + b^{l})$, where $W^{l}$ is the set of linear coefficients or \emph{weights} of layer $l$, $b^{l}$ is the set of independent terms or \emph{bias}, and $\sigma$ is the activation function. 

 {Here, we adopt a somewhat different approach and model how the data evolves through the network to establish a parallelism with dynamical systems. Accordingly, we define the following series of differential equations for any network property $Z$ evaluated locally at layer $l$ as}
\begin{equation}
	z^l=\frac{dZ}{dl} = g^l(Z,\phi^{l}), \quad l=1, \ldots, L.
\label{Eq:DiffLearning}
\end{equation}
 {Note that $Z$ is independent of the network parameters, 
but depends on the local transformation $\phi^{l}$ to which the data is subjected. The form of the function $g^l$, that depends on these transformations $\phi^{l}$, must be set for each type of layer, and are defined in Sec.~\ref{Sec:LocalMetricLayers}.} 

 {Subsequently, we define the cumulative property $Z^{L_{N}}$ in a layer $l=L_{N}$ as the sum of the transformations applied from the first to the current layer after going through all the previous layers
as}
\begin{equation}
	Z^{L_{N}}=\int_{1}^{L_{N}} g^l(Z,\phi^{l}) \hspace{3pt} dl.
	\label{Eq:Cumulative}
\end{equation}
 {Again, this is a scalar value that defines the cumulative evolution of the data at any point of the network. Finally, we obtain the \emph{global cumulative property} of the network, $Z_{GC}$, when it is evaluated in the output layer ($l=L$). A different global metric is also proposed, called the \emph{global sum property} 
$Z_{GS}$. It is estimated as the direct sum of all of the local property values, but without taking into account the topology of the architecture. Mathematically,} 
\begin{equation}
	Z_{GS} = \sum_{l=1}^{L} z^{l}.
	\label{Eq:GlobalSum}
\end{equation}

The proposed global metrics are easily estimated in a classical feedforward architecture, such as the multi-layer perceptron, by adding the local values consecutively. However, parallel paths and \emph{shortcut} (or \emph{skip}) connections often appear in many modern architectures. Hence, the network topology must be taken into account. That is why, we define the concept of \emph{layer algebra} to estimate the global metric values by considering all the information paths (see Sec.~\ref{Sec:LayerAlgebra}).  {In the end, we obtain unique values of the capacity and compression of a network to model how the data flows through the network regardless of the learned (optimized) values of the network parameters.}

\section{Local Properties Estimation}
\label{Sec:LocalMetricLayers}

Now that we have introduced the properties, { in this section, we first present their general definitions before specifying them} for some of the most common layers in a neural network. These properties are local and their contributions should be properly combined to obtain the global values (see Sec.~\ref{Sec:LayerAlgebra}).

\subsection{{ Generic Definitions}}

{ Although this section defines the presented metrics in a practical way for the most used operations (convolution, pooling, fully connected layers, etc.), it is not possible to define them for all the existing operations. For this reason, we define the general theoretical framework so that these metrics can be extrapolated to other types of operations.}

\begin{figure*}[!h]
	\centering
	\includegraphics[width=\textwidth]{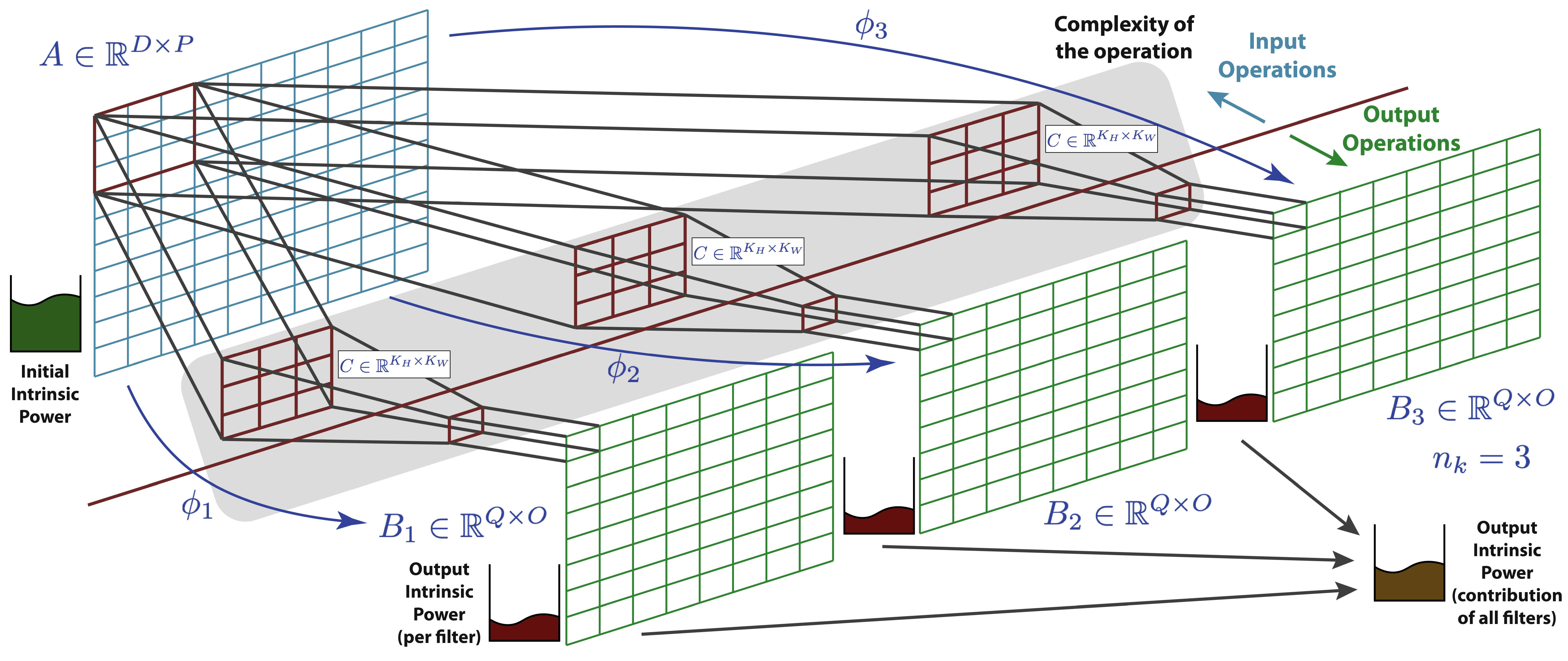}
	\caption{ 2D Convolution operation involving several filters illustrating local intrinsic power and complexity metrics.}
	\label{FigConvExplained}
\end{figure*}

{ Let $A \in \mathbb{R}^{D\times P}$ be a tensor object at the input of a local operation (input data) (see Fig.~\ref{FigConvExplained}) and $B \in \mathbb{R}^{Q \times O}$ be the tensor object obtained after applying the operation (output data). As defined earlier, $\phi^l$ is the mapping (transformation) between the input and output such that $B = \phi^l(A)$ and $\phi^l: \mathbb{R}^{D \times P} \mapsto \mathbb{R}^{Q \times O}$. Any local metric, as applied to $\phi^l$, is then defined in two different ways, depending on whether the operation involves weights (kernel-type operation) or is an activation function.}

{ If the (tensor) operation involves weighting an input signal (convolution, transpose convolution, pooling, or fully connected, among others), the intrinsic power, $p^l$, is defined as the ratio of the output space dimension, $\Phi_{out}$, with respect to the input space dimension, $\Phi_{in}$. Mathematically,}
\begin{equation}
	p^l = \frac{\Phi_{out}}{\Phi_{in}}.
	\label{EqGenericIPKernel}
\end{equation}
{ Here, the size of the output (input) space is analogous to the volume of data flow, which can be correlated to the flow energy, and, therefore, power. The denominator term acts as a normalizing factor so that the ratio captures the local data compression due to $\phi^l$. } 

{ If the operation is an activation function, the value of the intrinsic power is defined as the ratio between the variance of the output data with respect to the input data, where}
\begin{equation}
    p^l=\frac{\sigma^2_{out}}{\sigma^2_{in}}.
    \label{EqGenericIPActivation}
\end{equation}
 {This definition resembles the physical relationship between the transport power and amplitude of a wave, where power is proportional to the square of the amplitude. In our case, network data flow is analogous to wave transport and amplitude is captured through the data (sample) variance.}

{ 
Some other techniques could be applied, like the well-known K-L divergence, to measure the differences between the input and output distributions. However, the activation functions apply abrupt changes to the input data, 
causing problems in estimating the differences between the two distributions
as they are defined over very different ranges. Therefore, we decided to use the ratio of the outputs and inputs variances as a measure of the intrinsic power of an activation function. Further, since we are not evaluating the transitory learning process, we assume that the addition of the bias term would produce a readjustment of the classifier to obtain the same accuracy. Hence, it is not included in $p^{l}$ computation.
}

 {The second property to estimate is the capacity of the convolution operation using the complexity metric. As in the case of intrinsic power, we make a distinction between kernel-based operations and activation functions, and define complexity differently in the two cases.}

     {If the operation involves weighting the input signal, as in the case of convolution, transpose convolution, pooling, and fully connected layer, we define complexity, $c^l$, as the logarithm of the size of the local (weighting) tensor operator. Therefore,}
        \begin{equation}
    c^l=\log_2 (\Phi_{local}), 
    \label{EqGenericCWeighting}
    \end{equation}
     {where $\Phi_{local}$ is the size (dimension) of $\phi^l$. 
    This definition ties the layer complexity to a scaled (logarithmic) form of the local capacity (expressivity) of the corresponding operator in a simple manner. Note that the scaling, which preserves monotonicity, is needed to ensure that the complexity values are $O(1)$, just like for all the other ratio-based local property estimates.}
    
     {If the operation is an activation function, we simply interpret complexity as the inverse of its intrinsic power. 
    Therefore,}
    \begin{equation}
    c^l=\frac{1}{p^l}=\frac{\sigma^2_{in}}{\sigma^2_{out}}.
    \label{EqGenericCActivation}
\end{equation}
 {This interpretation is based on the notion that the capacity of a non-linear activation function can be estimated by the relative reduction in output variability due to the selection (high weighting) of specific input neurons. }

 {As in the case of intrinsic power, we only consider the changes that imply differences in the distribution of the data, which ultimately translates into modifying the classification capacity of the network. For example, some linear transformations, such as bias addition, 
do not have any effect in the decision capacity of a classifier and are, therefore, not taken into account.}

\subsection{Specific Definition: Convolution}
\begin{figure}[h]
	\centering
	\small{
	\includegraphics[width=0.49\textwidth]{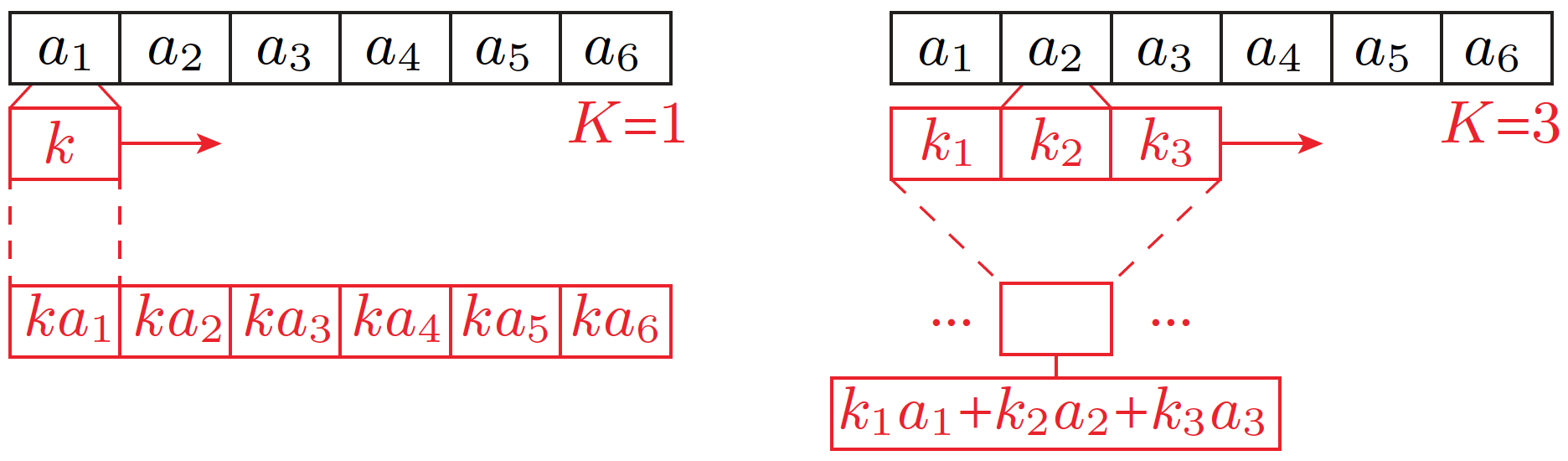}}
	\caption{Intrinsic power explanation in 1D convolutions using different kernel sizes.}
	\label{fig1DConvolution}
\end{figure}

We start with a simple 1D convolution example to understand our proposed properties and then apply it to 2D convolutions. The process can be generalized to higher dimensions. Let us assume that a kernel of size $K=1$ and a stride of $S=1$ is used in a 1D convolution. After applying the convolution to the data vector, the result is the same vector multiplied by the kernel element $k$. In other words, it is a linear combination of the initial values. This does not affect to the classification process, since the separation margins of the classifier would be adapted by $k$, resulting in the same accuracy. Hence, we assume that the intrinsic power of our data does not change, and $p^{l} = 1$ (see Fig.~\ref{fig1DConvolution}, left).

However, if a different kernel size of $K=3$
with a stride of $S=1$ are used, we can interpret the output as a \emph{softening} of the input values by means of a smoothing function with three coefficients (Fig.~\ref{fig1DConvolution}, right). This operation reduces the variability in the data. The output of (Fig.~\ref{fig1DConvolution}, right) is computed as the product of the kernel weights ($k_i, i=1,\ldots,K$) and the input vector over the cell $a_i$  and its $K-1$ (two) neighbors. This can be considered as the intrinsic power of that cell divided by $K$, as a result of applying the smoothness function. { If we extrapolate this operation to the rest of the input data, we can estimate a variation of the power of the data, defined as the particularization of Eq.~(\ref{EqGenericIPKernel}), giving the following equation}
\begin{align}
	p^{l}_{k} = \frac{\Phi_{out,k}}{\Phi_{in,k}} =\frac{C_o \cdot S_o}{K \cdot S_i}.
	\label{Eq:ConvLocalIntrinPower}
\end{align}
 {Correspondingly, complexity is given by}
\begin{align}
    c^{l}_{k}=\log_2 (\Phi_{local,k})=\log_2 K.
    \label{Eq:ConvLocalCapacity}
\end{align}
{ Here, $C_o$ is the number of output connections in the filter (typically equal to 1), $S_o$ is the size of the output data ($Q \times O$ in Fig.~\ref{FigConvExplained}),
$K$ is the size of the kernel ($K_H \times K_W$), and $S_i$ is the size of the input data ($D \times P$). Note that $S_o$ is related to the stride and depends on the size of the input data and \emph{padding}.} 

In the same convolution layer, it is common to apply several filters at the same time, obtaining a 3$^{rd}$ order tensor. For example, we could apply $n_{k}$ filters in the current layer, so that the output after this layer would be a block of size $n_{k}$ in the third dimension. Since we assume that the filters are completely independent with values other than zero, 
the intrinsic power of the data is multiplied by the number of filters. In other words, we are applying a smoothing function with each filter so that the power is reduced, but do it in parallel on all the filters so that the powers are added together. Therefore, the intrinsic power is written as
\begin{align}
	p^{l} = n_{k} \cdot p^{l}_{k}.
	\label{Eq:StdConvLocalIntrinPower}
\end{align}

The local complexity must also be multiplied by the number of filters in the layer since the individual filter capacities have to be added up as well. Therefore,
\begin{align}
	c^{l} = n_{k} \cdot c^{l}_{k}.
	\label{Eq:StdConvLocalCapacity}
\end{align}

\subsection{  Specific Definition: Transpose Convolution}

{  The intrinsic power of the transpose convolution is estimated using Eq.~(\ref{Eq:ConvLocalIntrinPower}) and Eq.~(\ref{Eq:StdConvLocalIntrinPower}), with the difference that $C_o$ is equal to $K$ and the number of input connections, $C_i$ is typically equal to 1. The expressions for complexity are the same as that for standard convolution since the local capacity is still equal to the size of the convolution filter.}
\begin{align}
	p^{l}_{k} = \frac{\Phi_{out,k}}{\Phi_{in,k}} =\frac{K \cdot S_o}{C_i \cdot S_i}.
	\label{Eq:TransConvLocalIntrinPower}
\end{align}
\begin{align}
    p^l=n_k \cdot p^{l}_{k}
\end{align}
\begin{align}
    c^{l}_{k}= \log_2 K; \quad c^l = n_k \cdot c^{l}_{k}
\end{align}

\subsection{Specific Definition: Pooling}
Pooling is an operation that can be viewed as a specific form of convolution. For this reason, we apply the same equations to estimate the intrinsic power (Eq.~(\ref{Eq:ConvLocalIntrinPower})) and capacity (Eq.~(\ref{Eq:ConvLocalCapacity})) of the local layers. Although there are different types of pooling operations (i.e., \emph{Average Pooling} or \emph{Max Pooling}, among others), in this work, we apply the same equations based on the changes in the dimensions of the data to estimate the properties of these type of layers.

However, there is a type of pooling operation that is worth analyzing separately. This operation is known as \emph{Global Pooling}, where the data is compressed along the \emph{width} and \emph{height} of the input matrix (in the case of 2D operations). Based on our analysis, we interpret this operation as the application of a single global filter of the same size as the input data ($D \times P$). Hence, we continue using Eq.~(\ref{Eq:ConvLocalIntrinPower}) and Eq.~(\ref{Eq:ConvLocalCapacity}), but modify the filter term suitably.

\subsection{Specific Definition: Fully Connected Layer}
For this type of layer, we make use of Eq.~(\ref{EqGenericIPKernel}) to define the local intrinsic power. Here, the output size is simply the number of neurons in the output layer, $D_{out}$, and the input size is the number of neurons in the input layer, $D_{in}$. Therefore, 
\begin{align}
	p^{l} = \frac{D_{out}}{D_{in}}.
\end{align}
 {Correspondingly, the local complexity is given by a particularization of Eq.~(\ref{EqGenericCWeighting}) as
\begin{align}
	c^{l} = \log_2 (D_{out} \cdot D_{in}),
\end{align}
since the local operator is defined as a flattened weight matrix in $\mathbb{R}^{D_{out} \times D_{in}}$.}

\subsection{Specific Definition: ReLU Activation}
As we are not interested in the weight values, we assume that a ReLU activation function \cite{Glorot2011} removes half of the activations. This is an approximation, since the weights assume more positive values as the learning progresses, leading to non-zero centered distributions. However, since we cannot estimate these variations for each network and data set, we consider the intrinsic power value to be around 0.5. 

Data-driven analysis leads us to the same conclusion (see the Supplementary Material), where we estimate the intrinsic power value for both normal and uniform data distributions. For this purpose, we apply the relation between the output-input variances to compare the data distribution before and after the activation layer. We directly consider the output-input relation as the local intrinsic power value, $p^{l}$ = $0.584$, and its inverse, $c^{l}$ = $1.713$, as the complexity of the ReLU activation function.

We can find other activation functions that are variants of ReLU (e.g., ELU, LeakyReLU \cite{He2015}, swish \cite{Ramachandran2017}, etc.), some of which contain parameters that can be optimized. This fact means that it is challenging to estimate the properties for these types of activation functions, since they depend on the data used and the learning step itself.
Consequently, we approximate the intrinsic power and complexity with the same values used for the ReLU function.

\subsection{Specific Definition: TanH Activation}

Since this type of activation function is highly dependent on the input values, it is more difficult to estimate a theoretical value for our metrics. Therefore, we rely more on the data-driven analysis, setting a value of $p^{l}$ = $0.628$ for the local intrinsic power (see the Supplementary Material). The value of the local complexity is again estimated as the inverse of the local intrinsic power to be $c^{l}$ = $1.592$.

It is worth pointing out here that the \emph{gain} of the activation function, related to the initialization values of the activations \cite{He2015,Glorot2010}, should not be confused with the idea proposed in our work. This \emph{gain} is related to the stability of the activation distributions in deep networks, seeking convergence of the output values after crossing a large number of layers.

\subsection{Specific Definition: Sigmoid Activation}

Similar to the previous case, it is also difficult to obtain theoretical values of intrinsic power and complexity for this type of activation function. We, therefore, again make use of data-driven analysis to set the values of $p^{l}$ = $0.208$ and $c^{l}$ = $4.802$.

\subsection{Specific Definition: Softmax Function}

This type of function is often used in classification problems to obtain an output vector that estimates the probability of belonging to a specific class. The input to a softmax function is not bounded, while its output lies within the interval (0,1) such that the sum of all its components is equal to 1. We can, therefore, understand that this function normalizes the data to a posterior probability distribution \cite{Gao2017}.

The softmax activation function changes the data distribution substantially by scaling and centering the data using exponential functions. Hence, theoretical values of the intrinsic power and complexity are especially hard to estimate. That is why we again rely on the data-driven approach to obtain a local intrinsic power value of $p^{l}$ = $1.342e$-$05$ and a local complexity value of $c^{l}$ = $7.454e04$.

\subsection{Specific Definition: Batch Normalization}
Batch normalization is a procedure to improve the learning process and stabilize the activations along the network \cite{Ioffe:2015aa}. It can be interpreted as a statistical adjustment that centers and scales the range of the activation values. This implies that the data is not changed as far as the prediction accuracy of a classifier is concerned. Batch normalization implies a movement in the limits of the classifier, sometimes making the values more suitable for the activation functions, but without modifying the relative internal variability of the data \cite{Santurkar:2018aa}. Therefore, even though it is proven to improve the training process \cite{Ioffe:2015aa}, it does not modify the capacity of the network. It improves the training process from the point of view of data adequacy and optimization, which, as already mentioned, is outside the scope of this work. Therefore, we conclude that batch normalization does not modify the intrinsic power of the data nor adds complexity to the network.

\subsection{Specific Definition: DropOut}

Dropout is a technique to increase the effectiveness of the training process in deep networks by reducing overfitting through \emph{co-adaptation} \cite{Srivastava2014}. It consists of dropping out some random units during the training step. However, the network recovers the complete set of neurons during the inference step. This implies that the overall capacity of the network remains unchanged, meaning that both the intrinsic power and network complexity are unaffected. 

\section{Layer Algebra}
\label{Sec:LayerAlgebra}
In the previous section, we defined the local intrinsic power and complexity values for some of the most common layers of a neural network. However, to estimate the global properties we have to take into account how the information flows throughout the network. The goal of this section is to develop this process with the definition of a simple concept that we call \emph{layer algebra}, understood as a set of simple operations to estimate the global properties of the network from the local values.

\begin{figure}[!h]
	\centering
	\small{
	\includegraphics[width=0.3\textwidth]{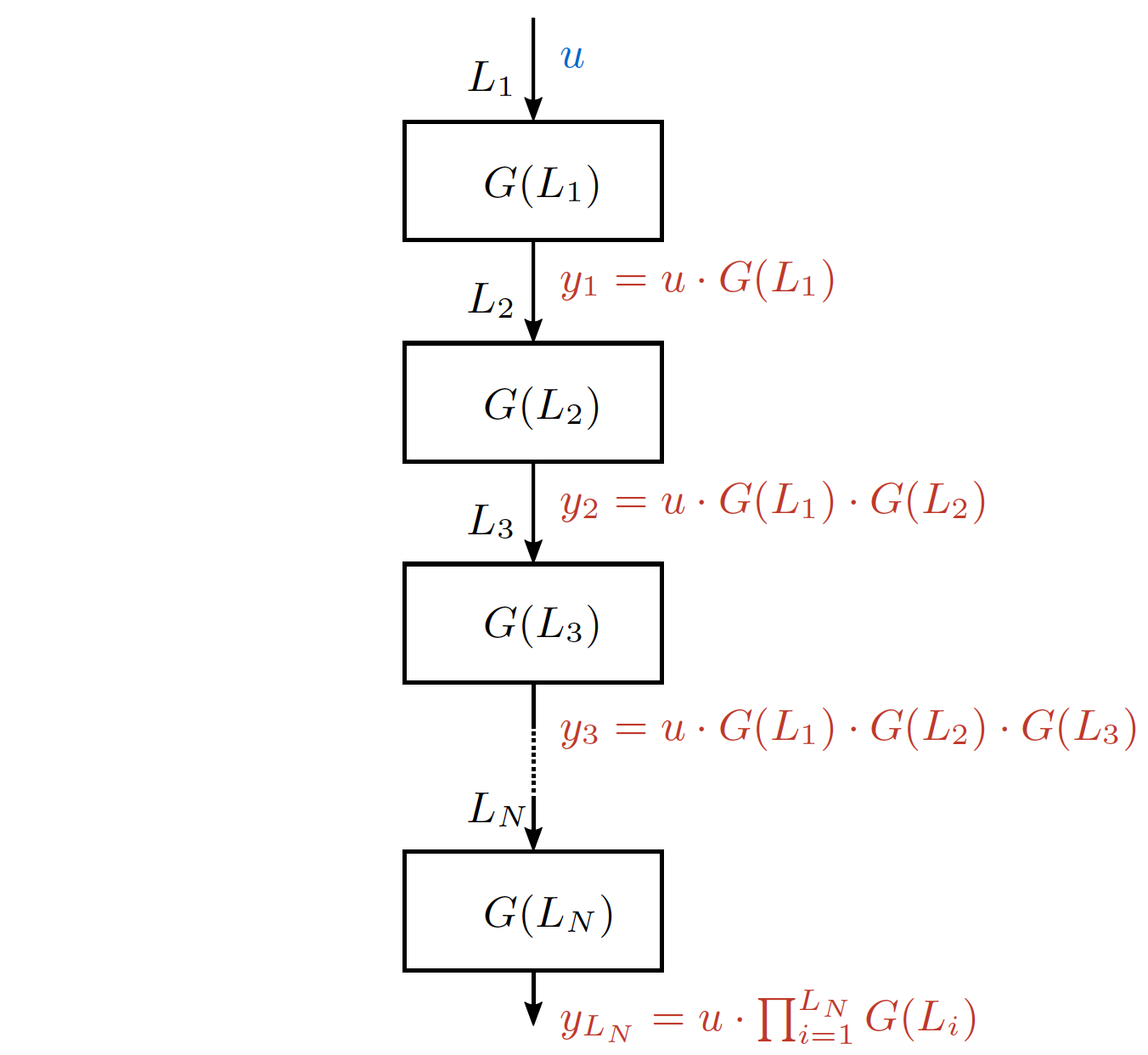}}
	\caption{Layer Algebra example to estimate the cumulative global properties of the network from layer (local) estimations. Functions $G(L_{n})$ represent the local metrics of our method (they can be \emph{Complexity} or \emph{Intrinsic Power}) and $y_{n}$ represent the cumulative values of these metrics.}
	\label{fig:TransfFunctions}
\end{figure}

To better understand the idea, let us consider a control theoretic point of view, where a neural network is defined as $\mathcal{A}(X,\theta)$ and the contribution of each network layer $l$ is interpreted as a modification of the original signal $u$ introduced to the network. Through simple operations, such as additions and multiplications, the \emph{amount of information} evolves as it flows through the network. In this way, each layer can be understood as a local transfer function $G(L_{i})$ that modifies the input signal $u$ until finally obtaining the output value $y_{L_{N}}=u \cdot \prod_{i=1}^{L_{N}}G(L_{i})$ in the layer $l = L_N$. Of course, the geometry of the neural network defines the operations to be carried out. For example, when the data passes through a layer, the process can be interpreted as the product of the input signal and the local function of the layer (see Fig.~\ref{fig:TransfFunctions}). In architectures with parallel channels or shortcut connections, as, for example, in the residual blocks of ResNets \cite{He:2016aa}, the contribution of both the parallel signals of the block must be taken into account (see Fig.~\ref{fig:TransfFunctionsParallel}).

\subsection{Intrinsic Power}
To estimate the cumulative value of the intrinsic power of the data, we assume that the network entry power has a value $ u = 1 $. Note here the parallelism with control theory. The fact that the input value is unitary instead of encoding the dimensions of the training data (tensor for images) is because they are considered in the first transfer function ($G(L_{1})$ in Fig.~\ref{fig:TransfFunctions}), corresponding to the first layer of the network. This implies   that we are moving all the data transformations to the transfer functions estimated in any local layer, allowing us to measure the intermediate cumulative values of the intrinsic power of the data at any point of the network ($y_{n}$ in Fig.~\ref{fig:TransfFunctions}).

In simple feedforward architectures, like a multi-layer perceptron (without parallel connections), the procedure is shown in Fig.~\ref{fig:TransfFunctions}. The value of the transfer function in any layer is equal to the local intrinsic power value, i.e., $G(L_{i}) = p^{i}$. The result of the cumulative intrinsic power after passing through the first layer is computed as the product of the transfer function $G(L_{1})$ and the input intrinsic power, in this case $u=1$. Using the general nomenclature of layer algebra, we define it as
\begin{equation}
	y_{1} = u \cdot G(L_{1}).
\end{equation}
Now, specifying for intrinsic power using the nomenclature defined in Sec.~\ref{Sec:NetProperties}, the cumulative intrinsic power is expressed as
\begin{equation}
	P^{L_{i}} = P^{L_{i-1}} \cdot p^{i}.
\end{equation}
The cumulative intrinsic power value is computed successively across all the layers until the global value is obtained.

\begin{figure}[!h]
	\centering
	\small{
	\includegraphics[width=0.3\textwidth]{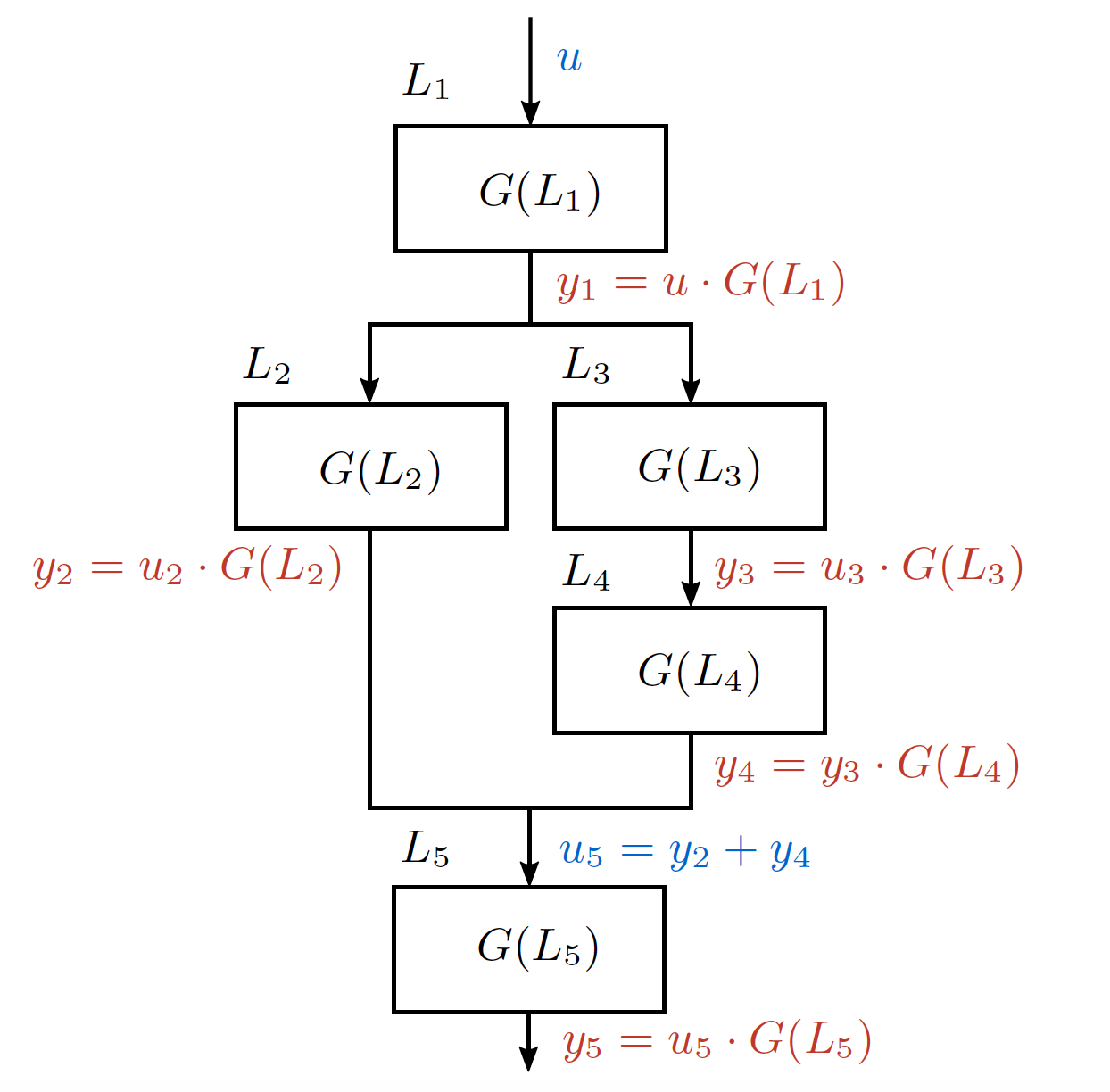}}
	\caption{Layer Algebra computation example with parallel connections. Again, functions $G(L_{n})$ represent the local metrics of our method (they can be \emph{Complexity} or \emph{Intrinsic Power}) and $y_{n}$ represent the cumulative values of these metrics.}
	\label{fig:TransfFunctionsParallel}
\end{figure}

In the case of an architecture with parallel paths, a slightly different analysis is necessary. For example, in the architecture shown in Fig.~\ref{fig:TransfFunctionsParallel}, a second branch appears in the output of layer $l = L_{1}$. According to our approach, this bifurcation implies that the power of the data is maintained in the input of the following layers, where $ y_{1} = u_{2} = u_{3} $. Continuing with the example of Fig.~\ref{fig:TransfFunctionsParallel}, the output information of layers $L_{2}$ and $L_{4}$ are received at the input of layer $l = L_{5}$. Since we are measuring the intrinsic power as a form of \emph{entropy} of the data, we consider the cumulative value of intrinsic power in the input of layer $L_{5}$ as the largest value of the previous outputs $u_{5} = \text{max}(y_{2},y_{4})$, since the value is stronger.

Although it is true that the activation values for the outputs of multiple converging layers are added (for example, in a residual ResNet block), their powers can be very different. Some of the paths can make deep transformations on the data, whereas, the others may copy the data without modifications. In other words, there is a mixing of information at different scales even though the data has the same size. For this reason, we believe that the highest value of intrinsic power should dominate the others.
Therefore, at a path convergence point, we have
\begin{equation}
	P^{L_{i-1}} = \text{max}(P^{L_{in,i}}),
	\label{Eq:InputIntrinPowerBifurc}
\end{equation}
where $P^{L_{i-1}}$ is the value of intrinsic power before layer $L_{i}$ and $P^{L_{in,i}}$ is the set of intrinsic power values of the paths entering in layer $L_{i}$. Through Eq.~(\ref{Eq:InputIntrinPowerBifurc}), we enforce consistency in the value of our metrics, since during the previous branching (bifurcation), we assumed duplication of the power by assuming a constant value of the intrinsic power, regardless of the number of paths that were created.

\subsection{Complexity}

In the case of complexity, again we first estimate the local values previously defined in Sec.~\ref{Sec:LocalMetricLayers}, depending on the type of layer. These values are represented by the transfer functions $G(L_{i})$ in Figs.~\ref{fig:TransfFunctions} and \ref{fig:TransfFunctionsParallel}, for the complexity.

The computation of the cumulative values of the complexity is carried out in the same way, through the product of a unit value of the input complexity with the local values of the layers that the data goes through. However, in the case of parallel paths (Fig.~\ref{fig:TransfFunctionsParallel}), the cumulative complexity is not computed as the maximum value of the inputs. Here, the cumulative complexity is estimated as the sum of all the values of the input paths for the layer $l=i$
\begin{equation}
	C^{L_{i-1}} = \sum_{j=1}^{N_{in}} (C^{L_{in,j}}),
\end{equation}
where $N_{in}$ is the number of inputs received by the layer $i$. In this case, we do not penalize the duplication of the complexity when branches appear, since we are measuring the capacity of the architecture, unlike intrinsic power that captures data compression.

\section{Experiments}

In this section we apply our metrics on four different experiments to demonstrate their usefulness. {  The first experiment is a comparison of a well-known metric (VC-Dimension) with our complexity metric.} The second experiment analyzes our metrics on a symmetric autoencoder. The third experiment compares the family of ResNets and their PlainNet counterparts using our metrics. Finally, section four gathers a set of widely-used state of the art architectures and tries to establish a correspondence between the complexity measure and the test accuracy of the architectures.

\subsection{{ Comparisons with VC Dimensions}}

\begin{figure}[!htbp]
	\centering		\includegraphics[width=0.459\textwidth]{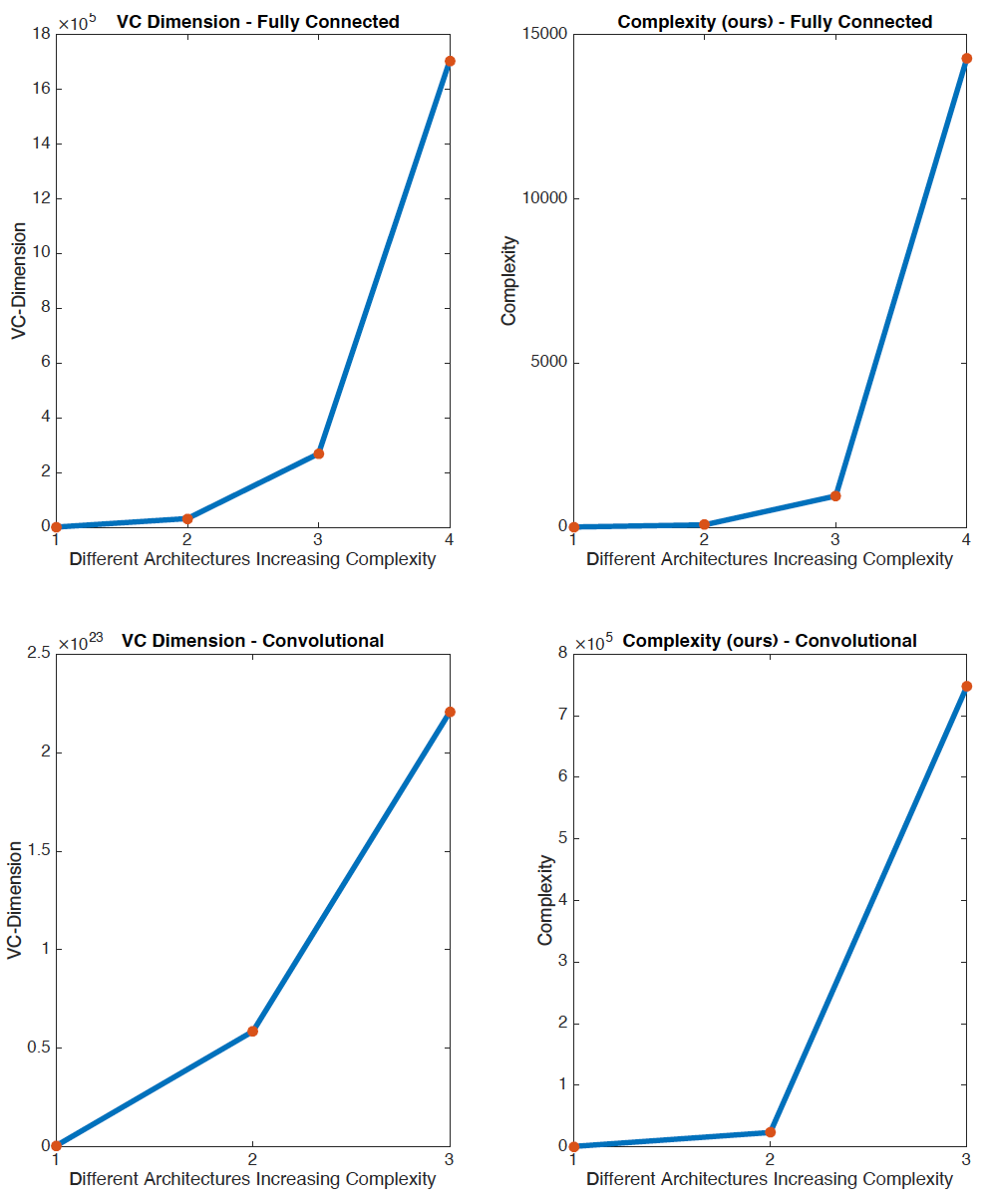}
	\caption{Comparison between VC-Dimension and our complexity metric for some examples of fully-connected and convolutional architectures.}
	\label{FigVCDimension}
\end{figure}

 {In this experiment, we consider a set of simple architectures and estimate their complexities using both our metric and the VC-Dimension (Vapnik–Chervonenkis). The VC-Dimension is a well-established way of estimating the complexity, and, therefore, the generalization error of certain classifiers. However, it can be quite challenging to estimate the VC-Dimension for deep networks comprising different types of layers. Recently, some theoretical estimates have been developed for fully-connected networks \cite{bartlett2019nearly} and convolutional networks with ReLU activation functions \cite{jiang2019fantastic}.  
Fig.~\ref{FigVCDimension} shows the results of our complexity metric and VC-Dimensions for these relatively simple architectures.}

 {Fig.~\ref{FigVCDimension} shows that the our complexity estimates follow the same monotonically increasing trends as the VC-Dimensions for all the architectures. However, the magnitudes of our estimates increase at substantially slower rates than VC-Dimensions for increasing network complexity. This implies that VC-Dimension can only be practically computed for simple architectures, whereas, our metric allows evaluation of very large or deep networks. Besides, our method allows flexibility in estimating the complexity of any convolutional or residual network comprising different types of layers. On the other hand, to the best of our knowledge, the theoretical approximations of VC-Dimension have not yet been developed for many layer types. Therefore, we cannot compare our complexity metric and VC-Dimension on the state-of-the-art architectures used in large-scale image classification problems.}

\subsection{Estimation of Properties for Autoencoder Network}

\begin{figure*}[!htbp]
	\centering
	\includegraphics[width=\textwidth]{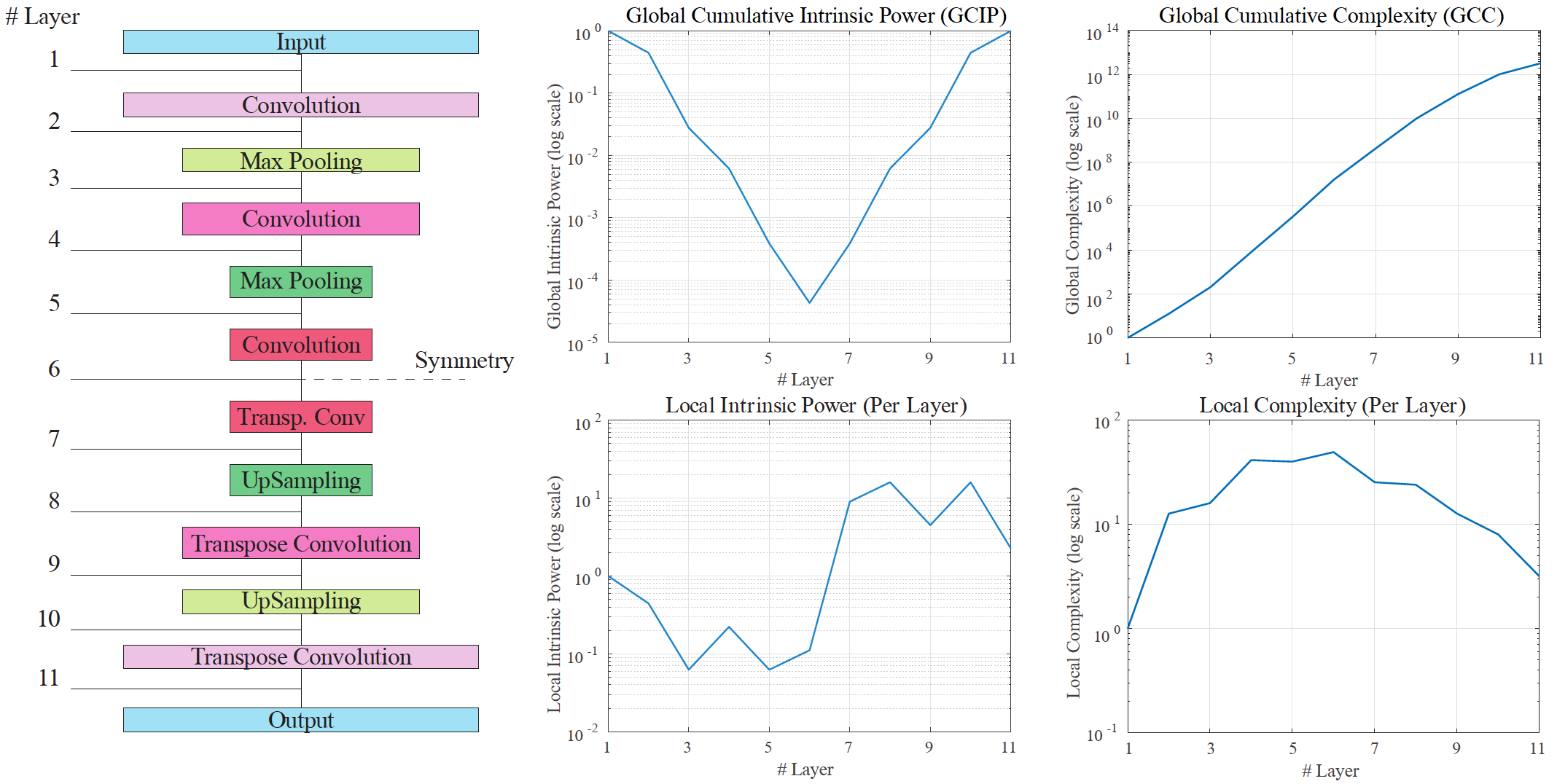}
	\caption{Layers of the proposed autoencoder and values of the intrinsic power and complexity of its architecture, both with local and global results.}
	\label{FigAutoencoder}
\end{figure*}

We now present a simple example to explain our metrics visually. It consists of a completely symmetric autoencoder with linear activation functions, where the arrangement of the layers is shown in the left section of Fig.~\ref{FigAutoencoder}. 
The intrinsic power plots are shown in the central section of Fig.~\ref{FigAutoencoder}, where the local values in each layer and the cumulative values are depicted in the lower and upper rows, respectively. Similarly, the layer-wise local and cumulative complexity values are shown in the bottom right and top right sections of Fig.~\ref{FigAutoencoder}, respectively. The global value of the metrics is taken as the cumulative value in the final layer of the network (GCIP and GCC for intrinsic power and complexity, respectively).

As seen in Fig.~\ref{FigAutoencoder}, the value of global intrinsic power is the same as the initial value, meaning that the power of the output data is the same as that of the input data. This is because we have used a completely symmetric architecture with linear activation functions in each layer. 
However, this does not imply that the value of the predicted labels would be the same as the input labels. 
It is also worth noting that a non-symmetric autoencoder with non-linear activation functions, or even a final classification function, is not likely to obtain the same output and input values of intrinsic power.

Although the power of the autoencoder output data reaches the same value as the input, the network does have a learning capacity, captured through a positive complexity value. Therefore, in the upper-right part of Fig.~\ref{FigAutoencoder}, we see the cumulative complexity of the network with an increasing trend throughout most of the local layers. {  The last value of the global cumulative complexity (GCC) is related to the expressivity of the autoencoder.}

\begin{figure*}[!htbp]
	\includegraphics[width=\textwidth]{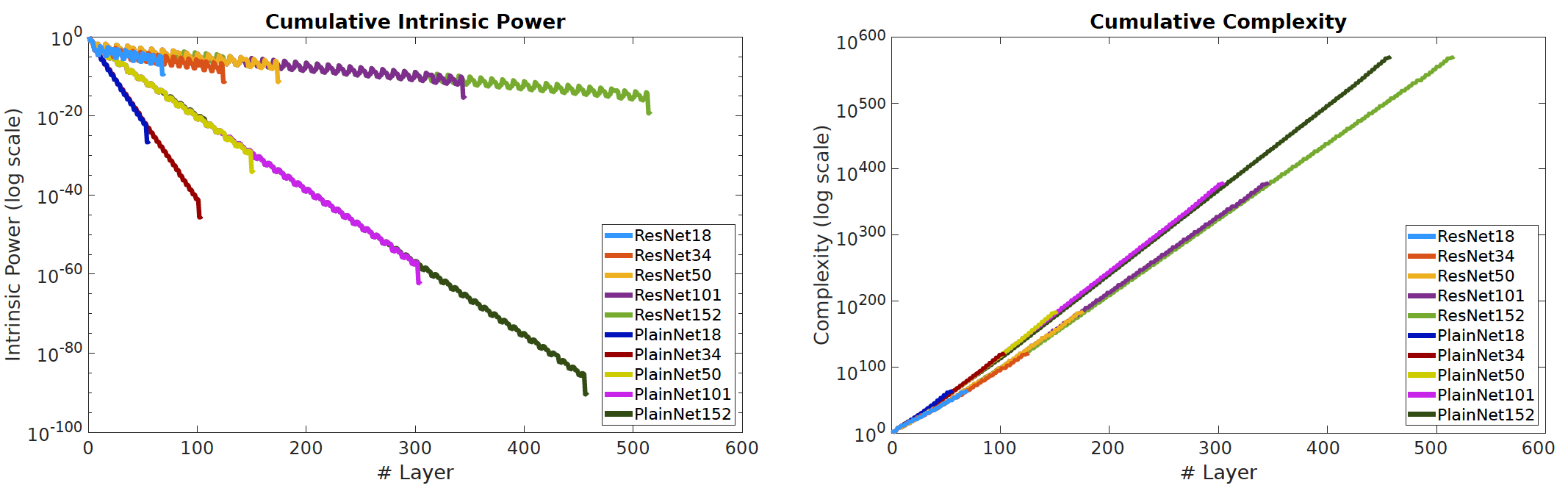}
	\caption{Comparison between the ResNet and PlainNet architectures with an equivalent number of layers, using the global cumulative intrinsic power (GCIP) and global cumulative complexity (GCC) metrics.}
	\label{fig:ResnetVsPlainNet}
\end{figure*}

\subsection{Comparisons Between ResNets and PlainNets}

After understanding how the metrics are computed and the relationship between the local and global values for a simple neural network, we now enumerate the contribution of our metrics. In this case, we measure the differences in the metrics values between the ResNet and the 
PlainNet family of architectures.

The results of the global metrics
for five models of the ResNet family and five models of the PlainNet family are shown in Fig.~\ref{fig:ResnetVsPlainNet}. Both the network families use the same number of processing blocks.
However, as shown along the horizontal-axis of Fig.~\ref{fig:ResnetVsPlainNet}, the number of equivalent local layers, corresponding to indivisible operations, are different. This happens as ResNet requires more operations to add the data in the residual blocks outputs, and certain operations, such as activations, occur independently along parallel network paths. 

In the right section of Fig.~\ref{fig:ResnetVsPlainNet}, we see very small differences between the global cumulative complexity values of the ResNet and PlainNet models. This means that the models have an equivalent capacity in that they can express or learn, essentially, the same set of functions for classification. Further, as might be expected, the complexity increases consistently with the number of layers, indicating that the deeper networks are capable of learning more complex patterns or structures from data.

Although the theoretical capacity of both the types of architectures is equivalent, their intrinsic powers are not. The left section of Fig.~\ref{fig:ResnetVsPlainNet} shows that there are notable differences between the ResNet family and the PlainNet family. These values can be understood as the power to compress data, where a small value indicates a large compression. Therefore, we can say that the PlainNet152 model has much higher compression than the ResNet152 model. In fact, all the models of the PlainNet family lead to a larger data compression than any member of the ResNet family. Within each family, of course, intrinsic power decreases, or, equivalently, data compression increases with network depth. Now, since the training process becomes more challenging with increasing data compression due to the difficulty in information flow across the network,
ResNets are easier to train than PlainNets as already pointed out in \cite{He:2016ab}.  {This observation is also indicated via both theoretical and experimental results in \cite{Rolnick:2018aa} (see Chapters 3.3.3 and 3.3.4), which further validates our choice of the intrinsic power metric and establishes its link to model learnability.} 

The fact that the learning process is improved in ResNet family as the data propagates more fluently is related to the final accuracy result. The fact that a network has the capability to express a classification function is a necessary but not sufficient condition. Through the training process, we must also ensure that the network is capable of learning the suitable weights to express the function. In other words, an architecture can be very complex (and, therefore, capable), but it may not be able to express its entire power if it is not trained adequately. Note though that a multitude of other factors play a role during the training process, such as the loss function, learning algorithm, batch normalization, data preparation, and data augmentation. Here, we only consider the intrinsic power to evaluate the learning capability of the network, as it as a very influential, data-independent factor in the training process.

It is interesting to note that the intrinsic power curve in the ResNet family of models produces small waves (Fig.~\ref{fig:ResnetVsPlainNet}), due to the summation of the powers from the shortcut channels. We also observe that the slope of the curves is steeper for the smaller models (with 18 and 34 layers) as compared to the larger models with a greater number of layers. This observation indicates that higher compression rates start happening earlier in small architectures, as might be conjectured intuitively.

\begin{figure}[h]
	\includegraphics[width=0.49\textwidth]{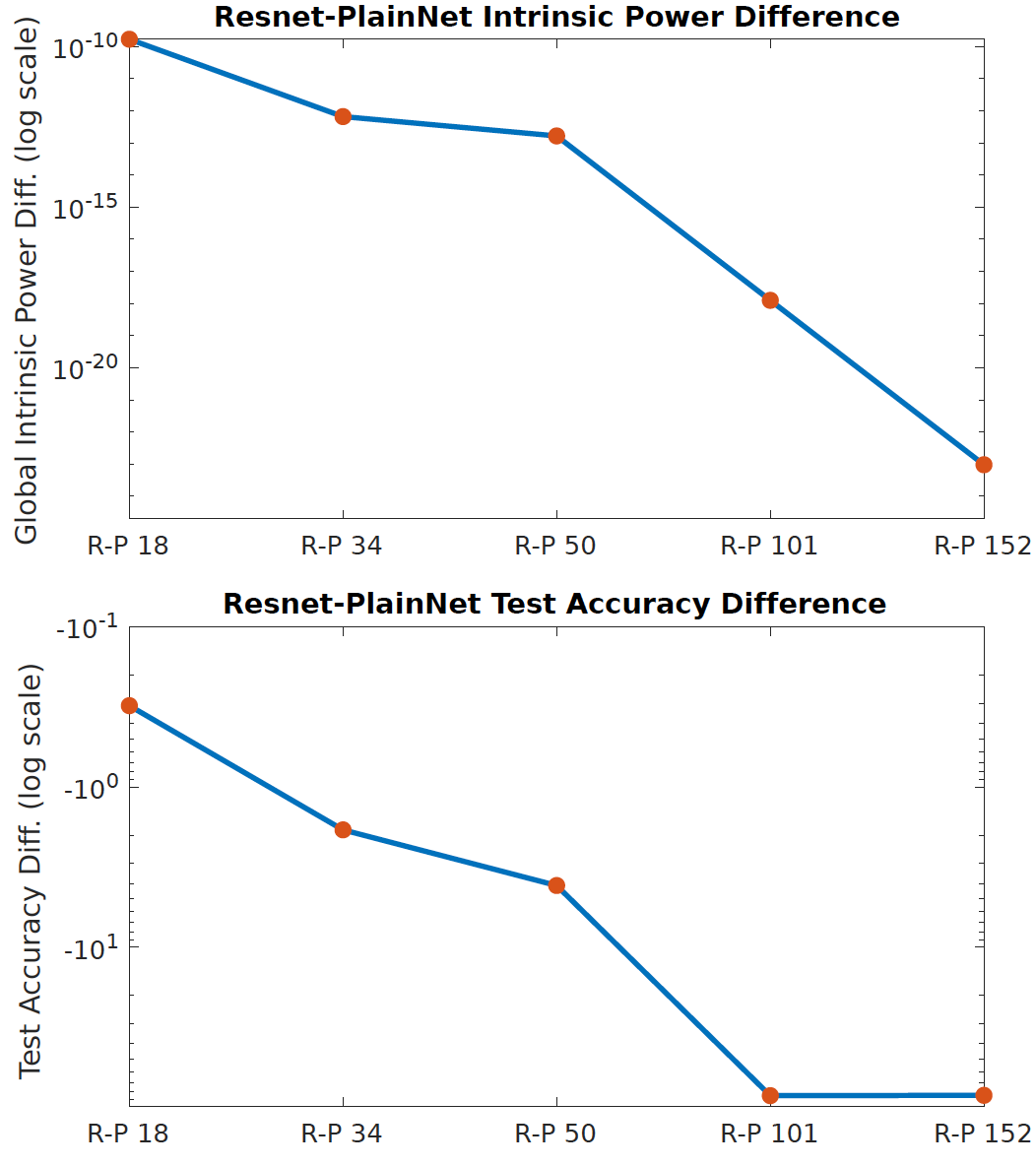}
	\caption{Comparison of ResNet and PlainNet architectures in terms of global cumulative intrinsic power and test accuracy on the CIFAR-10 dataset.}
	\label{fig:ResnetVsPlainNetCIFAR10}
\end{figure}

Fig.~\ref{fig:ResnetVsPlainNetCIFAR10} shows further analysis of the differences between the ResNet and PlainNet architectures for the same number of layers. The upper section of Fig.~\ref{fig:ResnetVsPlainNetCIFAR10} shows the differences in GCIP values, while the lower section shows the differences in test accuracy on the CIFAR-10 dataset \cite{Krizhevsky:2009aa}. These graphs indicate that, although the complexity of the ResNet and PlainNet models is nearly identical for the same number of layers (Fig.~\ref{fig:ResnetVsPlainNet}, right), test accuracy is substantially affected by the training process. As mentioned above, we believe the GCIP is a good estimator of the ease of training a model based solely on its architecture, and this is further supported from the similar trends observed in the two comparison graphs.
It should be noted that the PlainNet-101 and PlainNet-152 models are so hard to train that we could not achieve any meaningful test performance, due to which, the accuracy difference between the ResNet and PlainNet models for 101 and 152 layers is the absolute maximum. 

\subsection{Comparisons Among State of the Art Architectures}

\begin{figure*}[!htbp]
	\centerline{	\includegraphics[width=\textwidth]{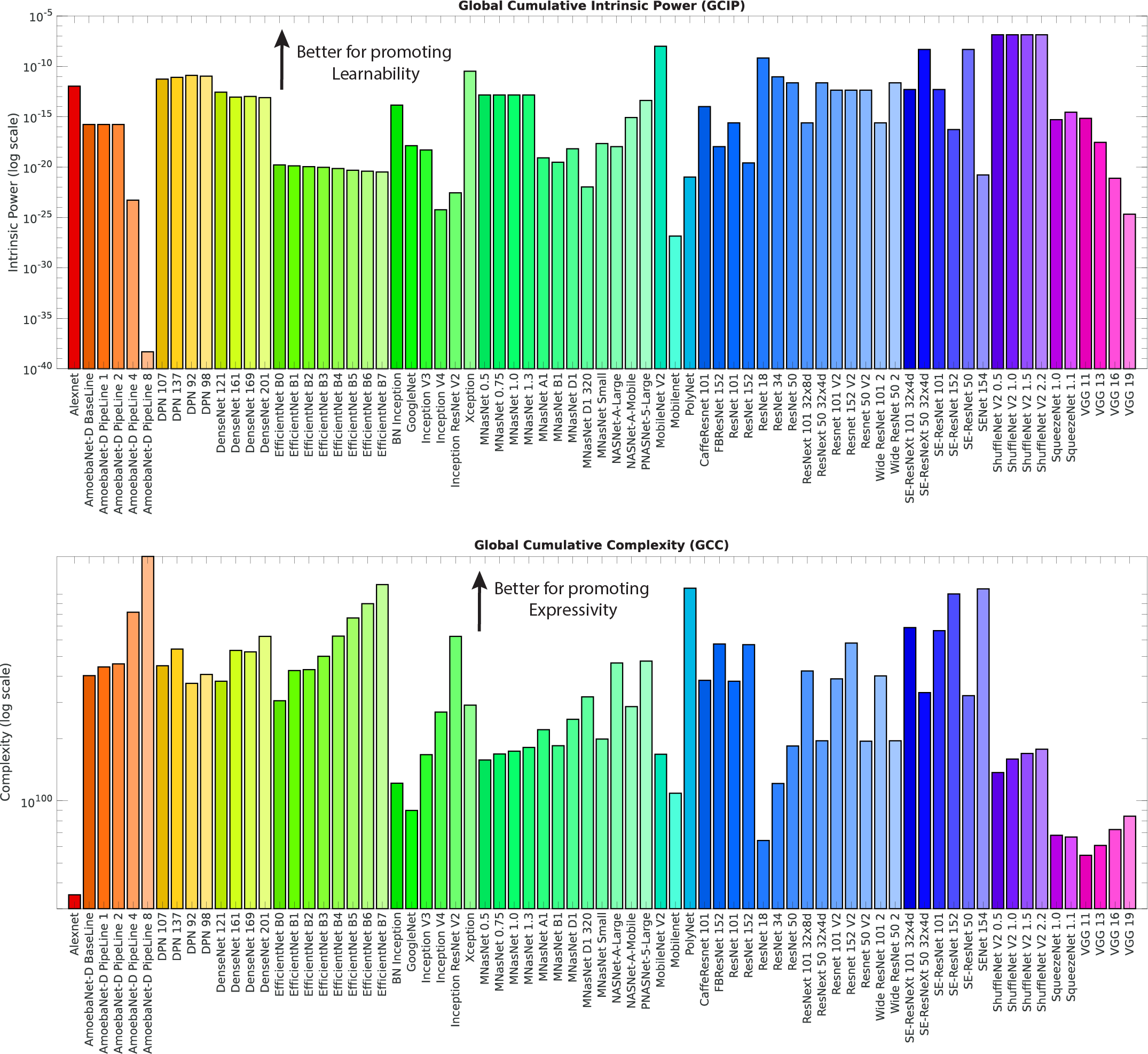}}
	\caption{Global cumulative intrinsic power (GCIP) and global cumulative complexity (GCC) values for some known network models grouped and colored by family. Low GCIP corresponds to a high transformation of the data, indicating that a moderately large value is desired to facilitate easy training. High GCC indicates that the network has a large capacity, implying that a high value is preferred, especially for challenging prediction problems.}
	\label{fig:ComplexityIntrinPowerAllArchs}
\end{figure*}

Finally, we test our metrics on a large set of deep neural networks. The network models are selected based on widespread use and easy access via open-source implementations in platforms such as TensorFlow \cite{Abadi:2016aa} and PyTorch \cite{Paszke:2019aa}. Fig.~\ref{fig:ComplexityIntrinPowerAllArchs} visualizes the GCIP and GCC values for all the models in the upper and lower section, respectively. This visualization allows us to quickly understand the variations of the metrics within the same family of networks as well as  compare the different architectures families.

\begin{figure*}[!h]
	\centerline{	\includegraphics[width=\textwidth]{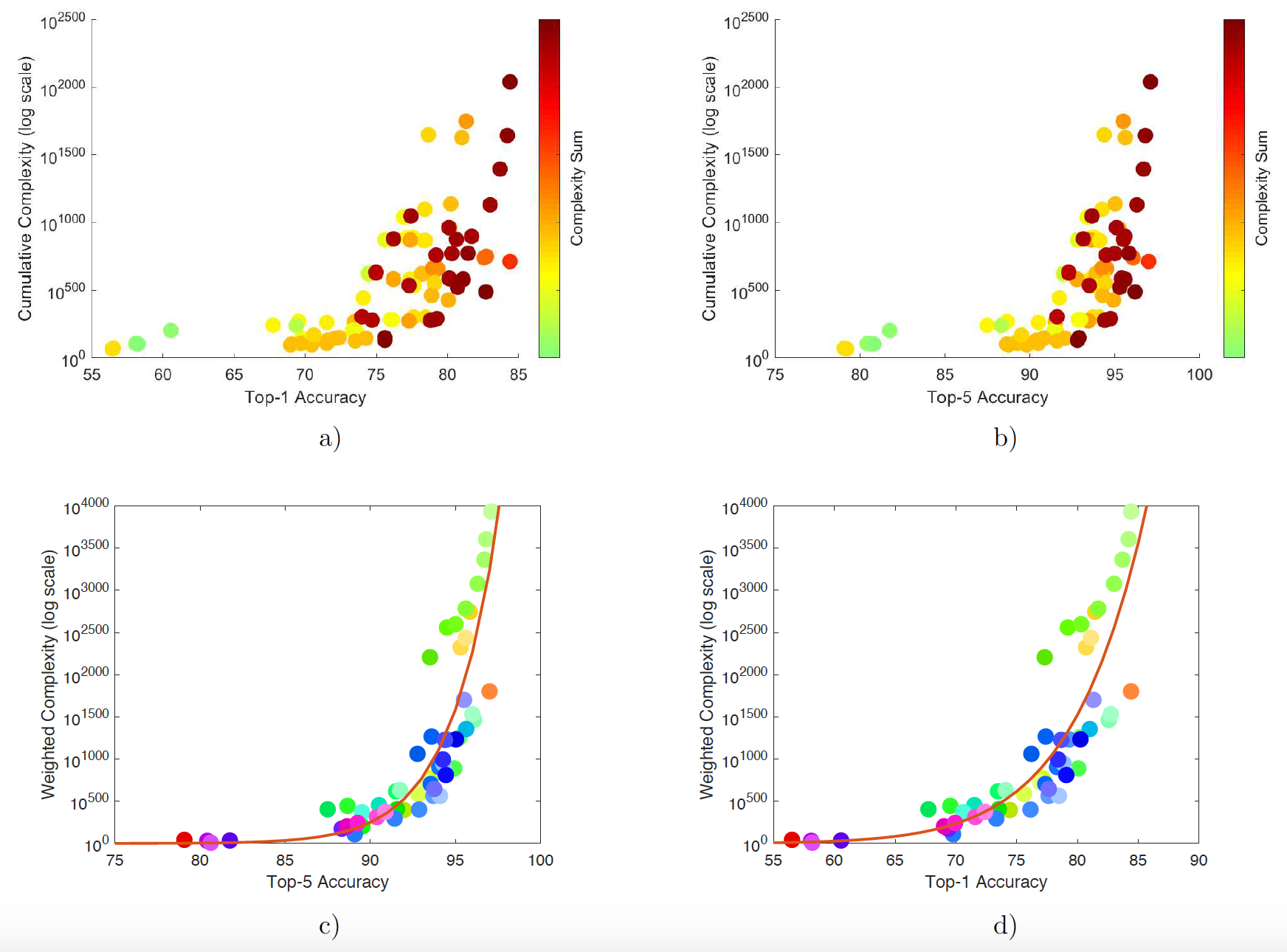}}
	\caption{Global complexity estimation with respect to (w.r.t.) ImageNet validation accuracy. a) Global cumulative complexity (GCC) w.r.t. Top-1 Accuracy with the color values representing the global sum complexity (GSC). b) GCC w.r.t. Top-5 Accuracy with the color values representing GSC. c) Global weighted complexity (GWC) w.r.t. Top-1 Accuracy with curve fitting and the same color scheme as in Fig.~\ref{fig:ComplexityIntrinPowerAllArchs} (more details are provided in the Supplementary Material). d) GWC w.r.t. Top-5 Accuracy with curve fitting.}
	\label{fig:CumulativeComplexity}
\end{figure*}

While a quantitative comparison of the complexity of the different network models is a contribution in itself, as in the previous experiment, we go a step further to investigate the correlation between our metrics and the accuracy of the models. To this end, we use a challenging benchmark classification dataset in the form of ImageNet \cite{Deng:2009aa}, and explain the variations in accuracy from a theoretical standpoint, for the first time to the best of our knowledge.

Specifically, we evaluate our global weighted and global sum complexity metrics with respect to the Top-1 and Top-5 accuracy in the ImageNet Large Scale Visual Recognition Challenge (ILSVRC) validation set. The results are shown in the upper section of Fig.~\ref{fig:CumulativeComplexity}, where we observe that both the complexity metrics are correlated to the accuracy achieved by the networks. 
However, the correlation shows a fair amount of dispersion, and we believe that a more useful relationship between complexity and accuracy can be obtained by reducing this dispersion. 
For this reason, we define a new variable called the \emph{global weighted complexity} (GWC), $C_{W}$, where the contributions of the global cumulative complexity (GCC) and the global sum complexity (GSC) are combined as
\begin{equation*}
	C_{W} = C^{L} * C_{GS},
\end{equation*}
with $l=L$ as the output layer of the neural network.

GWC yields a tighter relationship between network complexity and ImageNet validation accuracy, as shown in the bottom section of Fig.~\ref{fig:CumulativeComplexity}. We apply  curve fitting to obtain a numerical approximation of the relationship between the variables, where a power function gives correlation values of $R^2=0.751$ and $R^2=0.755$ for the Top-1 and Top-5 accuracy, respectively (orange curves in Fig.~\ref{fig:CumulativeComplexity}). This means that we are able to obtain a curve that suitably approximates the relation of the complexity of the networks with their classification accuracy (see the Supplementary Materials for more details). Therefore, 
our framework can not only describe the network properties and explain their predictive capabilities, it can also be potentially used to design new architectures for challenging learning tasks, by imposing a desired accuracy value on the fitted GWC curves, without requiring training.

\section{Conclusions}

It is still an open problem to estimate the inherent capacity of a (deep) neural network model to learn the target functions from the input to the output space. While the sum of parameters, number of layers, or memory size are often used in practice, we believe that they do not adequately capture the capacity of a model, and are, therefore, not suitable for comparing various models. Instead, we define a new layer-wise metric, called complexity, and combine the local values to come up with global cumulative complexity to encode the data-independent capacity of the network models. 
It is, however, important to note that the entire (theoretical) network capacity may not be available for any given dataset. In fact, capacity availability depends on several factors such as the ease and quality of the training process and the distribution of the input data. 

In addition to network complexity, we define a set of similar local and global intrinsic power metrics to estimate the compression applied on the data as they flow through the network. 
Deep compression implies that the data undergoes a big transformation during the inference process (direct propagation), and the gradients undergo large transformations during the learning process. In our opinion, this effect is closely related to the ease of learning. We understand that abrupt data transformations imply slower convergence, which translates into a more expensive learning process. Further, since we assign a specific value for each of these metrics, we are able to compare the different networks families and the different members within a family. This capability allows us to identify potent architectures with high complexities that are challenging to train. 

Furthermore, the slope of the cumulative curves shows the rate at which compression and complexity increase as the data passes through a given network, thereby offering hints on the desired architecture and depth to solve a given learning problem. Our proposed metrics also provide the possibility of estimating the image classification accuracy for any given neural network \textit{a priori}, since network complexity and prediction accuracy are found to be correlated. Therefore, we intend to use these metrics in the future to search for new architectures with the right balance of high complexity, relatively large intrinsic power (low compression), and reasonably low memory size (number of network parameters). 

 {We acknowledge that our framework does not consider all the contributing factors and performance measures of neural networks. In fact, we omit the effects of the initial values of the network weights, the learning algorithm and the corresponding hyperparameters (e.g., learning rate, batch size, use of momentum, etc.), and the optimization strategy in determining the prediction performance. Correspondingly, we do not include useful performance criteria such as robustness and optimization error. Future work would systematically investigate the trade-off between practical computability and a comprehensive \textit{a priori} analysis of networks. Such an investigation, along with further development of the generic property definitions, could also help in analyzing other types of networks, such as recurrent neural networks and transformers. }

\ifCLASSOPTIONcaptionsoff
  \newpage
\fi

\bibliographystyle{IEEEtran}
\bibliography{myBiblio}

\begin{IEEEbiography}[{\includegraphics[width=1in,height=1.25in,clip,keepaspectratio]{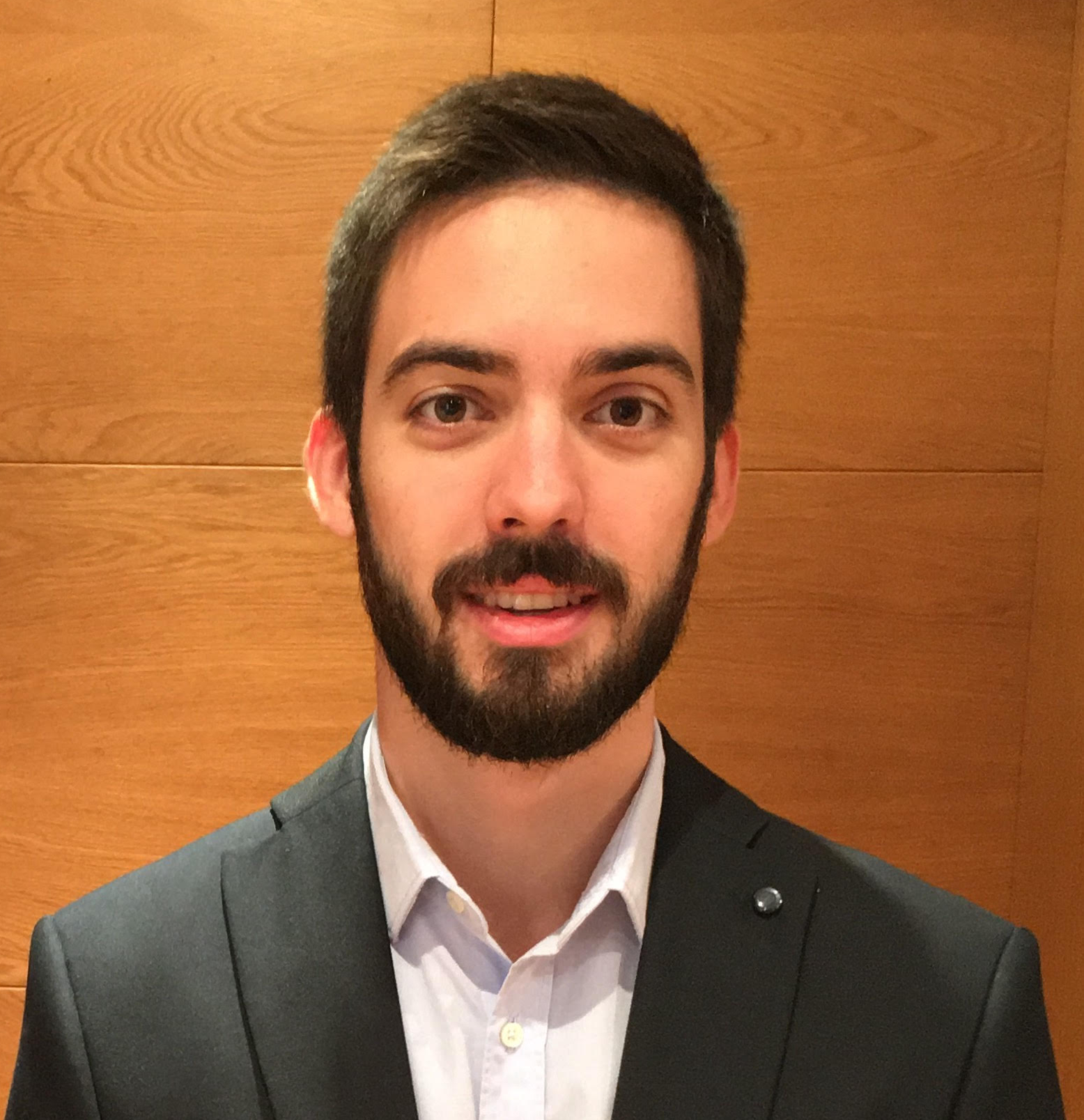}}]{Alberto Bad\'{i}as} is currently an Assistant Professor at the Polytechnic University of Madrid, Spain. He received the B.S. degree in mechanical engineering in 2011, the M.S. degree in industrial engineering (industrial automation and robotics) in 2014, the M.S. degree in biomedical engineering in 2016 and the Ph.D. degree in mechanical engineering in 2020, all from the University of Zaragoza. He has been working in the area of computer vision and robotics developing 3D reconstructions and new image processing tools, and is working currently on model order reduction methods and artificial intelligence in applied mechanics.
\end{IEEEbiography}

\begin{IEEEbiography}[{\includegraphics[width=1in,height=1.25in,clip,keepaspectratio]{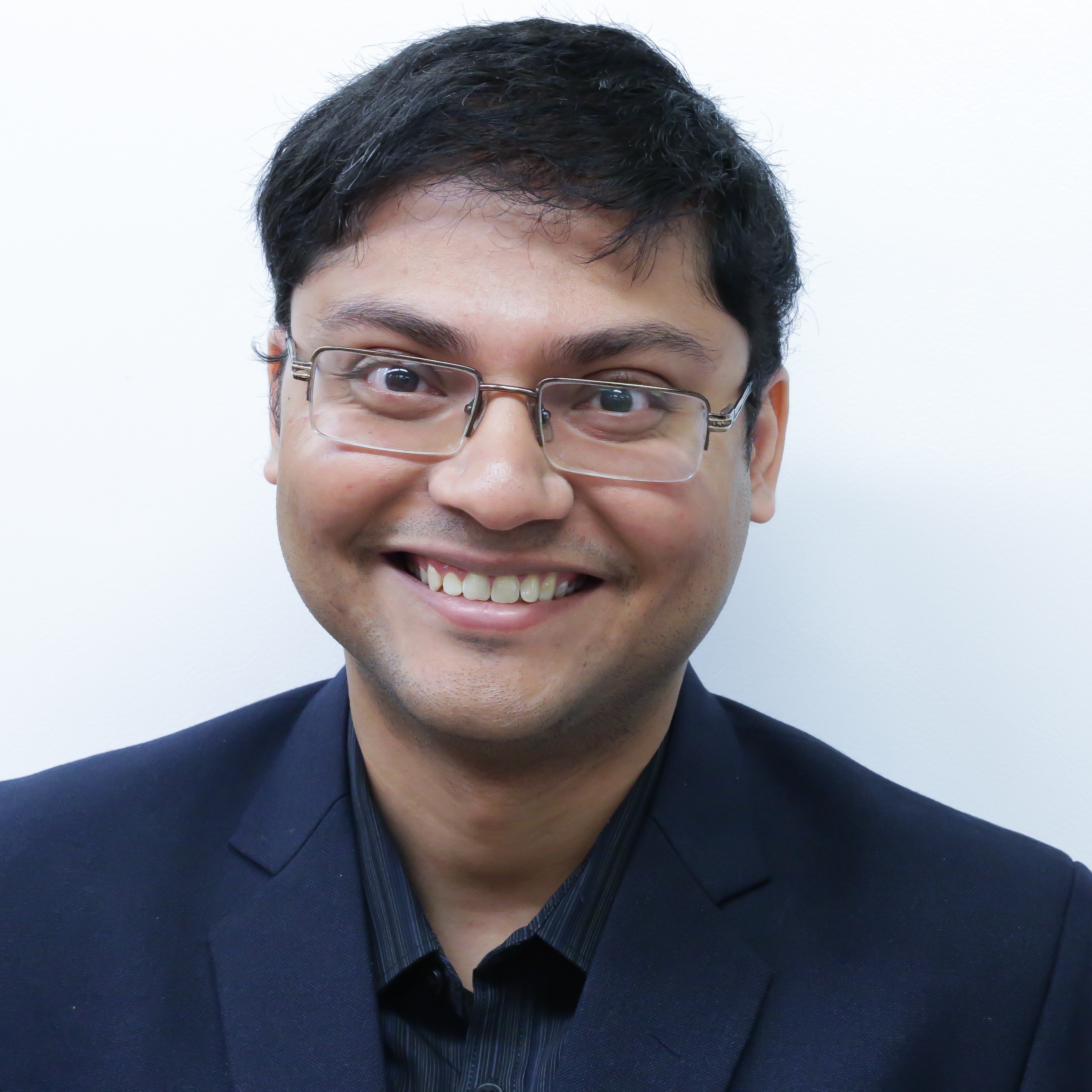}}]{Ashis G. Banerjee} (Senior Member, IEEE) received the
B.Tech. degree in manufacturing science and engineering from the Indian Institute of Technology Kharagpur, Kharagpur, India, in 2004, the M.S. degree in mechanical engineering from the University of Maryland (UMD), College Park, MD, USA, in 2006, and the Ph.D. degree in mechanical engineering from UMD in 2009.

He is currently an Associate Professor of industrial and systems engineering and mechanical engineering at the University of Washington, Seattle, WA, USA. Prior to this appointment, he was a Research Scientist at GE Global Research, Niskayuna, NY, USA. Previously, he was a Research Scientist and Post-Doctoral Associate at the Computer Science and Artificial Intelligence
Laboratory, Massachusetts Institute of Technology, Cambridge, MA, USA. His research interests include autonomous robotics, predictive analytics, and statistical learning.
\end{IEEEbiography}

\newpage
\appendices
\section{Experiments}
\label{Sec:AppExperiments}
In this supplementary material, we provide additional information on data-driven estimation of local intrinsic power and report detailed accuracy results for ImageNet classification.

\subsection{Data-Driven Estimation of Local Intrinsic Power for Various Network Layers}
\label{Supp:DataDriven}

\begin{figure}[!hbtp]
	\centering
	\includegraphics[width=0.49\textwidth]{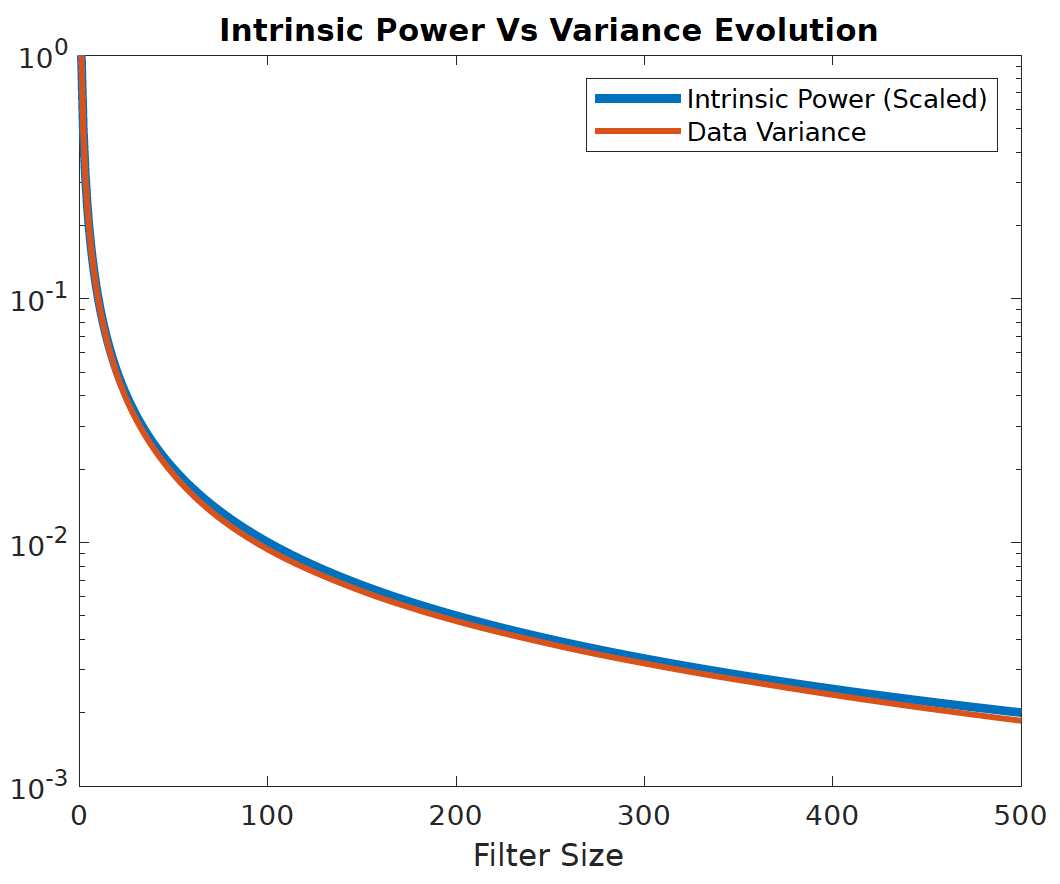}
	\caption{Intrinsic power and data variance evolution with respect to filter size.}
	\label{fig:IntrinPowerVariance}
\end{figure}

To check the validity of our intrinsic power measure empirically, we performed a data-driven study to estimate the data compression as it passes through a convolutional layer. We created $N_{V}$ = $100$ random vectors of size $[1 \times 15000]$, following a standard normal distribution $\mathcal{N}(0,1)$, and applying a filter of variable size $K=1,\ldots,500$, which is monotonically increasing (see Fig.~\ref{fig:IntrinPowerVariance}). To reduce the variability of the experiment, we forced the filter kernel values to $1/K$. The variance values shown in Fig.~\ref{fig:IntrinPowerVariance} are the average of the result over the $N_{V}$ vectors, while the values of the intrinsic power curve are scaled by the variance of the first random original vector. Fig.~\ref{fig:IntrinPowerVariance} shows that the curves are almost identical for a wide range of the filter size. This observation confirms that our calculation to estimate the variation of the local intrinsic power in convolutional layers is appropriate.

Figure 12 shows the output variances as functions of input variances for four different activation functions, when the input random variables are assumed to follow normal and uniform probability distributions.  

\begin{figure*}[!hbtp]
	\centering
	\includegraphics[width=0.75\textwidth]{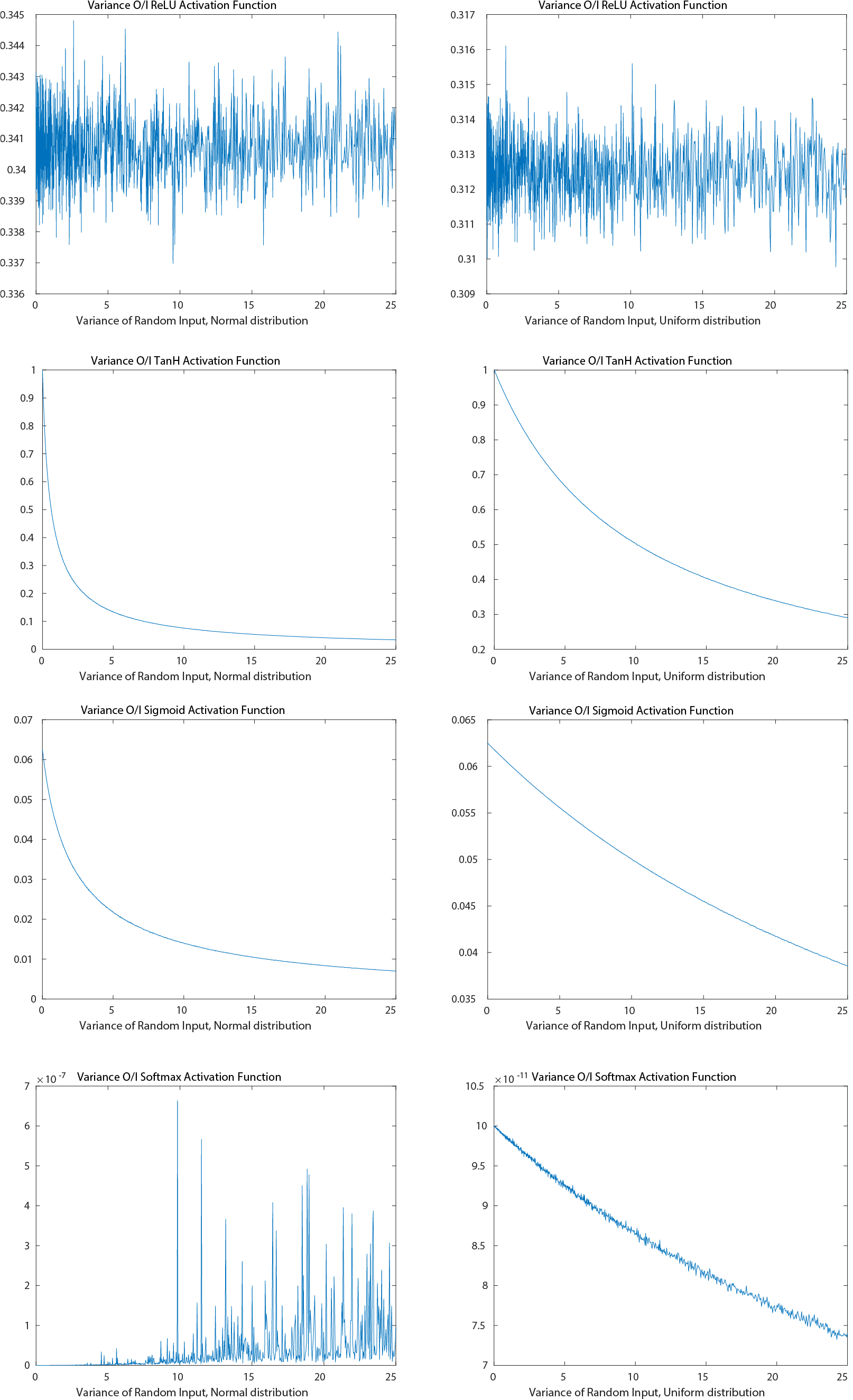}
	\caption{Output variances, as functions of input data variances, to estimate the data amplification of multivariate normal and uniform random variables using various activation functions.}
	\label{fig:AppVarOINormalUniform}
\end{figure*}

\subsection{ImageNet Classification Accuracy}
\label{Supp:ImageNet}

The following Tables I-III report the names and number of parameters of the state-of-the-art network models that are compared based on their ImageNet Top-1 and Top-5 accuracy values. Figures 13 and 14 supplement the global weighted complexity (GWC) curve fits w.r.t. the Top-1 and Top-5 accuracy, respectively, by illustrating the specific coordinates of all the evaluated deep network models. 

\begin{table}[!h]
\renewcommand{\arraystretch}{1.3}
\caption{PyTorch}
\label{table:Pytorch}
\centering
\begin{tabular}{|c|c|c|c|}
\hline
Model Name & Top-1 Acc. & Top-5 Acc. & N Params \\
\hline
AlexNet                 &    56.5180  &    79.0700  &      61100840  \\
DenseNet 121            &    74.4340  &    91.9720  &       7978856  \\
DenseNet 161            &    77.1380  &    93.5600  &      28681000  \\
DenseNet 169            &    75.6000  &    92.8060  &      14149480  \\
DenseNet 201            &    76.8960  &    93.3700  &      20013928  \\
GoogleNet               &    69.7780  &    89.5300  &       6624904  \\
Inception V3            &    69.5380  &    88.6540  &      23834568  \\
MNasNet 0.5             &    67.7340  &    87.4900  &       2218512  \\
MNasNet 0.75            &         NA  &         NA  &       3170208  \\
MNasNet 1.0             &    73.4560  &    91.5100  &       4383312  \\
MNasNet 1.3             &         NA  &         NA  &       6282256  \\
MobileNet V2            &    71.5140  &    90.5050  &       3504872  \\
ResNet 101              &    77.3470  &    93.5460  &      44549160  \\
ResNet 152              &    78.3120  &    94.0460  &      60192808  \\
ResNet 18               &    69.7580  &    89.0780  &      11689512  \\
ResNet 34               &    73.3140  &    91.4200  &      21797672  \\
ResNet 50               &    76.1300  &    92.8620  &      25557032  \\
ResNext 101 32x8d       &    79.3120  &    94.5260  &      88791336  \\
ResNext 50 32x4d        &    77.6180  &    93.6980  &      25028904  \\
ShuffleNet V2 0.5       &    60.5520  &    81.7460  &       1366792  \\
ShuffleNet V2 1.0       &    69.3620  &    88.3160  &       2278604  \\
ShuffleNet V2 1.5       &         NA  &         NA  &       3503624  \\
ShuffleNet V2 2.2       &         NA  &         NA  &       7393996  \\
SqueezeNet 1.0          &    58.0920  &    80.4200  &       1248424  \\
SqueezeNet 1.1          &    58.1780  &    80.6240  &       1235496  \\
VGG 11                  &    69.0200  &    88.6300  &     132863336  \\
VGG 13                  &    69.9300  &    89.2500  &     133047848  \\
VGG 16                  &    71.5900  &    90.3800  &     138357544  \\
VGG 19                  &    72.3800  &    90.8800  &     143667240  \\
Wide ResNet 101 2       &    78.8460  &    94.2840  &     126884696  \\
Wide ResNet 50 2        &    78.4680  &    94.0860  &      68883240  \\
AmoebaNet-D BaseLine    &         NA  &         NA  &      81505540  \\
AmoebaNet-D PipeLine 1  &         NA  &         NA  &     319024120  \\
AmoebaNet-D PipeLine 2  &    84.4000  &     97.000  &     542734840  \\
AmoebaNet-D PipeLine 4  &         NA  &         NA  &    1.0558e+09  \\
AmoebaNet-D PipeLine 8  &         NA  &         NA  &    1.8449e+09  \\
\hline
\end{tabular}
\end{table}

\begin{table}[!t]
\renewcommand{\arraystretch}{1.3}
\caption{PyTorch Repo 2}
\label{table:PytorchRepo2}
\centering
\begin{tabular}{|c|c|c|c|}
\hline
Model Name & Top-1 Acc. & Top-5 Acc. & N Params \\
\hline
AlexNet               &    56.4320  &    79.1940  &     61100840  \\
BN Inception          &    73.5240  &    91.5620  &     11295240  \\
CaffeResnet 101       &    76.2000  &    92.7660  &     44549160  \\
DenseNet 121          &    74.6460  &    92.1360  &      7978856  \\
DenseNet 161          &    77.5600  &    93.7980  &     28681000  \\
DenseNet 169          &    76.0260  &    92.9920  &     14149480  \\
DenseNet 201          &    77.1520  &    93.5480  &     20013928  \\
FBResNet 152          &    77.3860  &    93.5940  &     60268520  \\
Inception ResNet V2   &    80.1700  &    95.2340  &     55843464  \\
Inception V3          &    77.2940  &    93.4540  &     23834568  \\
Inception V4          &    80.0620  &    94.9260  &     42679816  \\
NASNet-A-Large        &    82.5660  &    96.0860  &     88753150  \\
NASNet-A-Mobile       &    74.0800  &    91.7400  &      5289978  \\
PNASNet-5-Large       &    82.7360  &    95.9920  &     86057668  \\
PolyNet               &    81.0020  &    95.6240  &     95366600  \\
ResNet 101            &    77.4380  &    93.6720  &     44549160  \\
ResNet 152            &    78.4280  &    94.1100  &     60192808  \\
ResNet 18             &    70.1420  &    89.2740  &     11689512  \\
ResNet 34             &    73.5540  &    91.4560  &     21797672  \\
ResNet 50             &    76.0020  &    92.9800  &     25557032  \\
ResNeXt 101 32x4d     &    78.1880  &    93.8860  &     44177704  \\
ResNeXt 101 64x4d     &    78.9560  &    94.2520  &     83455272  \\
SENet 154             &    81.3040  &    95.4980  &    115088984  \\
SE-ResNet 101         &    78.3960  &    94.2580  &     49326872  \\
SE-ResNet 152         &    78.6580  &    94.3740  &     66821848  \\
SE-ResNet 50          &    77.6360  &    93.7520  &     28088024  \\
SE-ResNeXt 101 32x4d  &    80.2360  &    95.0280  &     48955416  \\
SE-ResNeXt 50 32x4d   &    79.0760  &    94.4340  &     27559896  \\
SqueezeNet 1.0        &    58.1080  &    80.4280  &      1248424  \\
SqueezeNet 1.1        &    58.2500  &    80.8000  &      1235496  \\
VGG 11 BN             &    70.4520  &    89.8180  &    132868840  \\
VGG 11                &    68.9700  &    88.7460  &    132863336  \\
VGG 13 BN             &    71.5080  &    90.4940  &    133053736  \\
VGG 13                &    69.6620  &    89.2640  &    133047848  \\
VGG 16 BN             &    73.5180  &    91.6080  &    138365992  \\
VGG 16                &    71.6360  &    90.3540  &    138357544  \\
VGG 19 BN             &    74.2660  &    92.0660  &    143678248  \\
VGG 19                &    72.0800  &    90.8220  &    143667240  \\
Xception              &    78.8880  &    94.2920  &     22855952  \\
\hline
\end{tabular}
\end{table}

\begin{table}[!t]
\renewcommand{\arraystretch}{1.3}
\caption{TensorFlow}
\label{table:TFRepo2}
\centering
\begin{tabular}{|c|c|c|c|}
\hline
Model Name & Top-1 Acc. & Top-5 Acc. & N Params \\
\hline
NAS LargerNet        &    82.7000  &    96.2000  &     88949818  \\
NAS MobileNet        &         74  &    91.6000  &      5289978  \\
Densenet 121         &    74.9800  &    92.2900  &      7978856  \\
Densenet 169         &    76.2000  &    93.1500  &     14149480  \\
Densenet 201         &    77.4200  &    93.6600  &     20013928  \\
Inception Resnet V2  &    80.1000  &    95.1000  &     55813192  \\
Inception V3         &    78.8000  &    94.4000  &     23817352  \\
Mobilenet            &    70.6000  &    89.5000  &      4231976  \\
Mobilenet V2         &    74.7000  &        NaN  &      3504872  \\
Resnet 101           &    80.1300  &    95.4000  &     44601832  \\
Resnet 101 V2        &        NaN  &        NaN  &     44577896  \\
Resnet 152           &    80.6200  &    95.5100  &     60268520  \\
Resnet 152 V2        &        NaN  &        NaN  &     60236904  \\
Resnet 50            &    79.2600  &    94.7500  &     25583592  \\
Resnet 50 V2         &        NaN  &        NaN  &     25568360  \\
VGG 16               &    75.6000  &    92.8000  &    138357544  \\
VGG 19               &    75.6000  &    92.9000  &    143667240  \\
Xception             &         79  &    94.5000  &     22855952  \\
EfficientNet B0      &    77.3000  &    93.5000  &      5288548  \\
EfficientNet B1      &    79.2000  &    94.5000  &      7856239  \\
EfficientNet B2      &    80.3000  &         95  &      9109994  \\
EfficientNet B3      &    81.7000  &    95.6000  &     12233232  \\
EfficientNet B4      &         83  &    96.3000  &     19341616  \\
EfficientNet B5      &    83.7000  &    96.7000  &     30389784  \\
EfficientNet B6      &    84.2000  &    96.8000  &     43040704  \\
EfficientNet B7      &    84.4000  &    97.1000  &     66347960  \\
MNasNet A1           &        NaN  &        NaN  &      3887038  \\
MNasNet B1           &        NaN  &        NaN  &      4383312  \\
MNasNet Small        &        NaN  &        NaN  &      2030264  \\
MNasNet D1           &        NaN  &        NaN  &      3638404  \\
MNasNet D1 320       &        NaN  &        NaN  &      6932240  \\
DPN 92               &    80.7000  &    95.3000  &     37655904  \\
DPN 98               &    81.1000  &    95.6000  &     61553152  \\
DPN 107              &        NaN  &        NaN  &     86879216  \\
DPN 137              &    81.4500  &    95.8400  &     79221824  \\
\hline
\end{tabular}
\end{table}

\begin{figure*}[!hp]
	\centerline{
	    	\includegraphics[width=\textwidth]{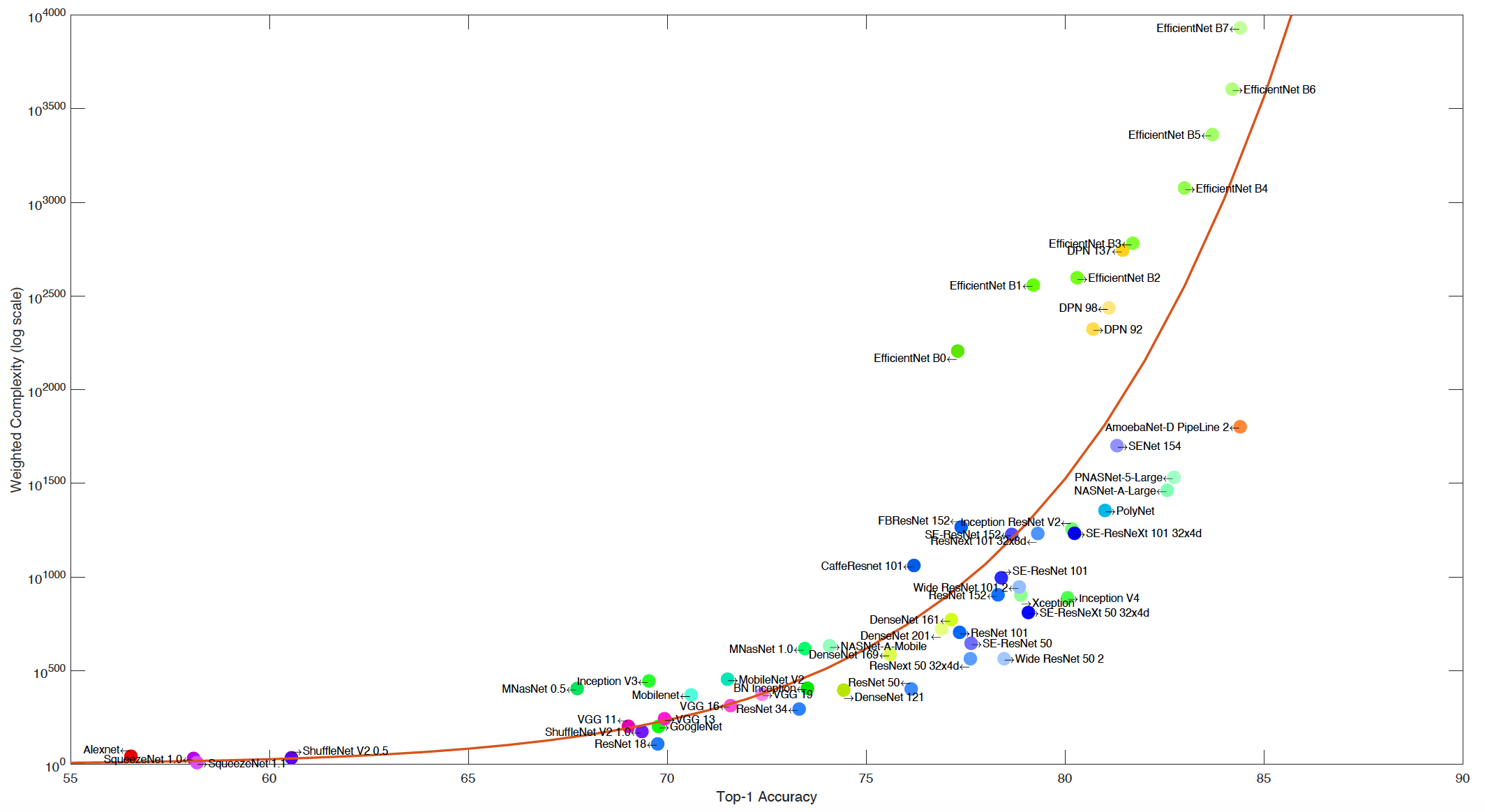}
	}
	\caption{Global Weighted Complexity (GWC) w.r.t Top-1 Accuracy with curve fitting. The color of each point of the graph is based on the scheme followed in Fig. 9.}
	\label{fig:App:GWC1PerFamily}
\end{figure*}

\begin{figure*}[!hp]
	\centerline{
	    	\includegraphics[width=\textwidth]{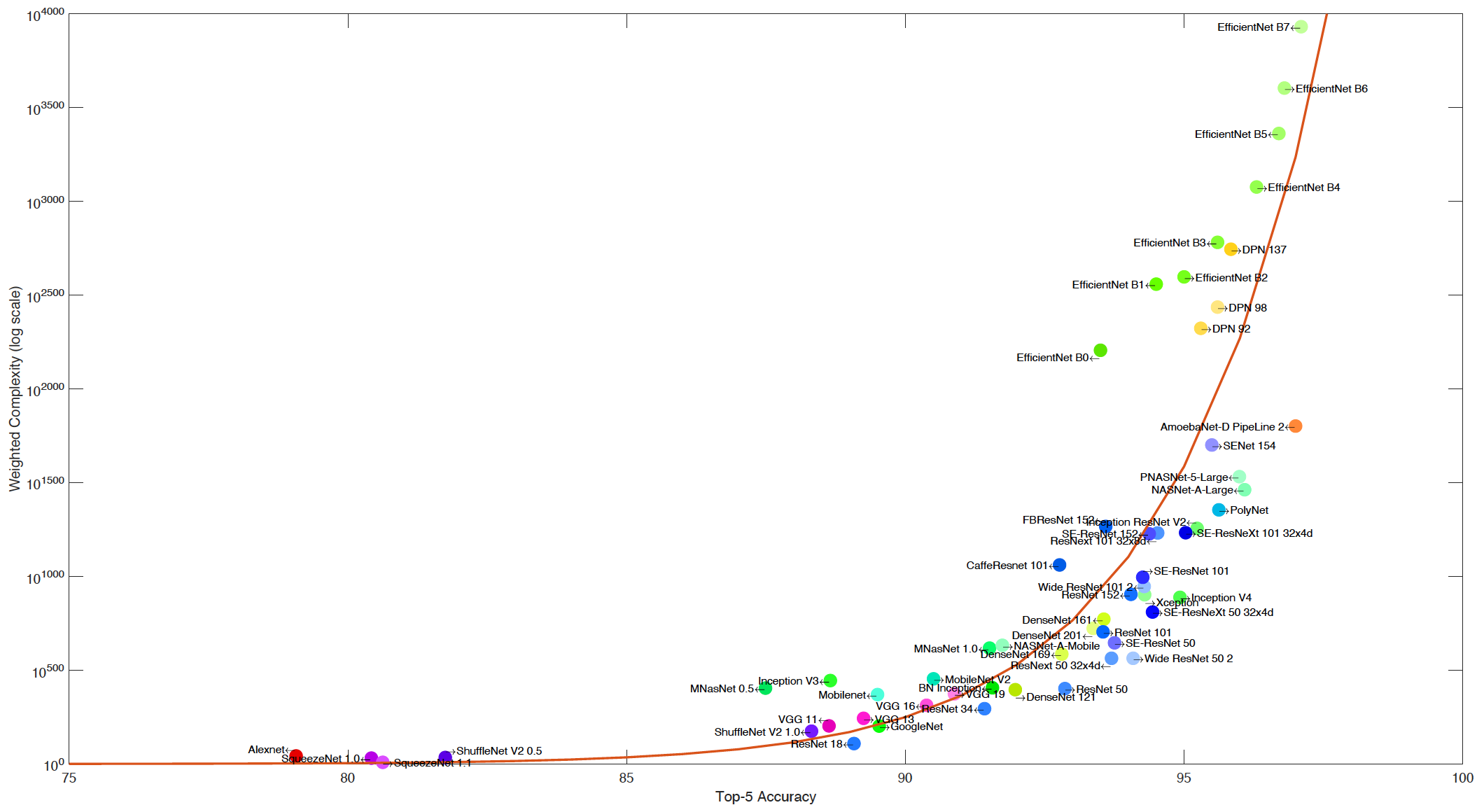}
	}
	\caption{Global Weighted Complexity (GWC) w.r.t Top-5 Accuracy with curve fitting. The color of each point of the graph is based on the scheme followed in Fig. 9.}
	\label{fig:App:GWC5PerFamily}
\end{figure*}

\ifCLASSOPTIONcaptionsoff
  \newpage
\fi

\end{document}